\documentclass{article}
\newif\ifshowappendix            
\showappendixtrue               
\PassOptionsToPackage{numbers, compress}{natbib}

\usepackage[preprint]{neurips_2025}

\usepackage[utf8]{inputenc} 
\usepackage[T1]{fontenc}    
\usepackage{hyperref}       
\usepackage{url}            
\usepackage{booktabs,tablefootnote}       
\usepackage{amsfonts}       
\usepackage{nicefrac}       
\usepackage{microtype}      
\usepackage{xcolor}         
\usepackage{graphicx}
\usepackage{subcaption}
\usepackage{amsmath}
\usepackage{amssymb}
\usepackage{algorithm}
\usepackage{algorithmic}
\usepackage{tabularx}
\usepackage{wrapfig}

\usepackage[capitalize]{cleveref}
\title{PLUMAGE: \\ Probabilistic Low-rank Unbiased Min-Variance \\ Gradient Estimator for Efficient Large Model Training}

\author{%
  Matan Haroush \\
  Department of Electrical Engineering\\
  Technion\\
  Haifa, Israel\\
  \texttt{matan.h@campus.technion.ac.il} \\
  \And
  Daniel Soudry \\
  Department of Electrical Engineering\\
  Technion\\  
  Haifa, Israel\\
  \texttt{daniel.soudry@gmail.com} \\
}

\makeatletter
\newcommand{\RTCOMMENT}[1]{\hfill\(\triangleright\)~#1}
\makeatother

\begin{document}
\maketitle

\begin{abstract}
Accelerator memory and networking constraints have emerged as dominant bottlenecks when training \emph{large language models} (LLMs) with billions of parameters. Existing low-rank gradient estimators such as \textsc{GaLoRE} and \textsc{FLORA} compress gradients and optimizer tensors by projecting weight‑gradients onto a rank‑$r$ subspace, enabling LLM training on consumer hardware. Yet, these methods are either biased or subject to high estimator variance. Moreover, the optimizer state based on the first \emph{and} second-moment estimates expressed in the previous subspace becomes misaligned whenever the projection is updated, leading to instabilities during training.
We propose \textbf{PLUMAGE}: \emph{\textbf{P}robabilistic \textbf{L}ow‑rank \textbf{U}nbiased \textbf{M}inimum‑v\textbf{A}riance \textbf{G}radient \textbf{E}stimator}. 
PLUMAGE is a drop-in replacement for existing low‑rank gradient estimators; it does not introduce new hyperparameters beyond the chosen rank $r$ and the update interval $\tau$. In addition, we resolve optimizer state misalignment issues to prevent spurious weight updates and enhance training stability.
We empirically demonstrate that PLUMAGE shrinks the full-rank optimization's gap over the pre-training evaluation loss by $33\%$ on average across models, and the average training loss across the GLUE benchmark by $28\%$ -- within a similar computational and memory footprint as GaloRE.
\end{abstract}

\section{Introduction}
\label{intro}
Natural language modeling has benefited from training large-language models (LLMs), which consist of billions of parameters on massive amounts of unsupervised data. This typically requires distributed training algorithms employing tens of thousands of interconnected high-end accelerators to speed up the training time. Moreover, some of the largest state-of-the-art models do not fit in a single device's memory. Meanwhile, the historical progress of scaling the capabilities of high-bandwidth memory and device connectivity lags behind progress in scaling the compute capabilities of modern accelerators \citep{Gholami_2024aiwall}.
Thus, significant engineering and research efforts aim to improve the efficiency of training and inference. For example, a variety of methods have been proposed, leveraging model parallel computing \citep{shoeybi2019megatronlm, Rajbhandari_2020deepspeed, Zhao_2023fsdp}, structured sparsity \citep{NEURIPS2021_mnsparsity,chmiel2022minimum}, and applications of low-precision numerical formats \cite{courbariaux2016binarized,banner2018scalable,chmiel2020neural,dettmers20218bit,blumenfeld2024cheaper}. 
 
As models grow in size, the memory and connectivity limitations of modern accelerators emerge as a critical bottleneck, impeding the ability of the research community to train and fine-tune state-of-the-art models without relying on access to expensive hardware and infrastructure. Addressing these constraints is essential for reducing entry barriers for future research.
One recent avenue of work aims to enable the training and fine-tuning of large models with limited accelerator memory overhead. 
For example, parameter-efficient fine-tuning methods inject trainable adapters into frozen large models when learning a new task. For instance, \citet{hu2021loralowrankadaptationlarge} (LoRA) proposed attaching low-rank parametrized adapters to the linear layers of a given model to augment its activations. This approach reduces optimizer state memory requirements compared to full model training, reduces the risk of overfitting on small datasets, and potentially incorporates compute and memory gains from low-bit numerical formats (e.g., \textsc{QLoRA}, \citet{dettmers2023qloraefficientfinetuningquantized}). However, these methods rely on having a pre-trained model and are less suited when one wishes to train the entire model.

Recently, \citet{galore} (\textsc{GaLoRE}) proposed projecting gradients to a low-rank subspace. The method projects the dense linear layers' weight gradients, reducing the optimizer's state memory without enforcing a low-rank weight structure, leading to a potentially better compression rate and lower terminal loss. More importantly, the authors showed how large language models can be trained from scratch on widely available consumer-grade hardware with limited memory. Notably, the gradient projection at the core of the method is constructed by stacking the top-$k$ singular vectors of the weight gradient that are periodically computed via Singular Value Decomposition (\textsc{SVD}, \citet{eckart1936approximation}).

These results led us to question whether there is a better low-rank gradient estimation alternative for training neural networks, given that the gradients are not necessarily of low rank. Moreover, fixing the projection to the span of the top-$k$ singular vectors can lead to a significant accumulation of bias during the optimization process, which is known to be detrimental \citep{chmiel2020neural,chen2023symbolic}.
Our contributions can be summarized as follows:
\begin{itemize}
\item We derive a novel $k$-sparse \textbf{P}robabilistic \textbf{L}ow-rank \textbf{U}nbiased \textbf{M}inimum-v\textbf{A}riance \textbf{G}radient \textbf{E}stimator (\textsc{PLUMAGE}). Our approach relies on an efficient and fixed-rank sampling strategy without replacement. In addition, it requires permanent storage of only a one-sided projection matrix per weight, similarly to \citet{galore}.
\item We develop an alignment method for the first and second moments used by stateful optimizers such as \textsc{Adam}. Our alignment strategy mitigates the adverse effects of periodic projection subspace updates during training and enhances the training stability with low-rank gradient estimators.
\item We empirically demonstrate a significant improvement in convergence when using low-rank gradient estimators --- without tuning the full-rank learning rate or adding hyperparameters other than the rank of choice and projection update interval.


\end{itemize}
\section{Related Work}
\label{related}
This section contains foundational and closely related work necessary to contextualize our approach.

\textbf{Momentum and Adaptive Optimizers}.
Given learning rate $\eta$ and the gradient $\mathbf{G}_t=\nabla \mathcal{L}_t$ at step $t$, vanilla stochastic gradient descent (\textsc{SGD}) updates the weight matrix in each layer according to 
\begin{equation}
{\mathbf{W}}_{t+1} = \mathbf{W}_t - \eta \mathbf{G}_t,.
\end{equation}
Adding a momentum term with a $\beta_1$ coefficient to dampen stochastic gradient noise (\textsc{SGDM}, \citet{rumelhart1986learning}) yields
\begin{equation}
\label{eq:first_moment}
{\mathbf{M}}_t = \beta_1 {\mathbf{M}}_{t-1} + (1 - \beta_1) \mathbf{G}_t    
\end{equation}
\begin{equation}
\mathbf{W}_{t+1} = \mathbf{W}_t - \eta \mathbf{M}_t \,.
\end{equation}
In adaptive step-size optimizers, the learning rate is adjusted based on past gradients. For instance, \textsc{Adam} \cite{kingma2017adammethodstochasticoptimization} update rule is
\begin{equation}
\label{eq:sec_moment}
{\mathbf{V}}_t = \beta_2 {\mathbf{V}}_{t-1} + (1 - \beta_2) \mathbf{G}_t^{\circ 2}
\end{equation}
\begin{equation}
\label{eq:adam_update}
{\mathbf{W}}_{t+1} = {\mathbf{W}}_t - \frac{\eta {\mathbf{M}}_t}{\sqrt{{\mathbf{V}}_t} + \epsilon} \,.
\end{equation}
where ${\mathbf{V}}_t$ is the second-moment estimate of the gradients, and $\Box^{\circ 2}$ denotes element-wise square (bias-correction factors omitted).

\textbf{Memory-Efficient Optimizers}.
Despite its popularity and advantages, \textsc{Adam} incurs a high computational and memory overhead. Light-state optimizers such as Adafactor \cite{shazeer2018adafactor} and subsequent (e.g., Lion, Shampoo, etc., \citep{chen2023symbolic, gupta2018shampoo, vyas2024soap}), were introduced to mitigate this overhead. An alternative approach is to quantize the state optimizers (e.g., \citet{dettmers20218bit}). These methods are generally orthogonal to our work since they can be applied to a low-rank gradient as input.

\textbf{Sparse Gradients}. 
Sparse gradient methods such as coordinate descent focus on updating only a subset of model parameters, reducing compute, communication, and memory costs by dropping parts of the gradient (e.g., \cite{Aji_2017,chmiel2022minimum}). Recently, \citet{Grass-sparse-grad2024} proposed an efficient and light sparse gradient method dubbed GRASS. GRASS maintains row-sparse gradient statistics in its optimizer state. It is constructed as an unbiased and minimum variance gradient estimator by selecting rows according to their norm. However, the minimum variance property is under a multinomial sampling. The authors point out that analytically computing the minimum variance for a multinomial distribution without replacement is not tractable. Ultimately, allowing high norm rows to be sampled multiple times increases the estimator variance compared to non-replacement sampling. In addition, a row-sparse approach leads to instabilities during training. As a result, the authors resort to strategies such as momentum restarts and periodic learning-rate warm-ups when updating the sampled row indices.

\textbf{Low-Rank Gradients for Communication Compression.} 
In distributed settings, low-rank gradient methods project the gradients onto a lower-dimensional subspace when synchronizing the gradients between the workers before performing an optimization step. For example, \citet{wang2018atomo} (\textsc{ATOMO}) proposed a min-variance unbiased low-rank gradient estimator that reduces communication by sparsifying the gradient's singular values and transmitting only the surviving \textsc{SVD} components. \textsc{ATOMO}'s sampling strategy produces a rank-$k$ gradient \emph{on average} and requires communication of both the left and right projections. Later, PowerSGD \cite{vogels2020powersgd} leveraged the power-iteration method to maintain an approximation of the top-k gradient singular vectors, reducing the cost of applying SVD on the gradients before each communication, while adding a feedback mechanism to incorporate the previous step error in the next gradient communication.

\textbf{Low-Rank Gradients for Memory-Constrained Optimization.} 
\textsc{GaLoRE} and \textsc{FLORA} exploit the low-rank gradient structure to reduce device memory costs \citep{galore,hao2024flora} when training and finetuning large models that do not fit in the device memory.
Unlike previously mentioned \textsc{GaLoRE}, \textsc{FLORA} utilizes reconstructible random Gaussian projections (via random generator seed reuse) and is paired with \textsc{Adafactor} \citep{shazeer2018adafactor} optimizer, which maintains a factorized approximation of the second moment statistic used for the adaptive step-size. As such, it only maintains the projected gradient moment in memory. \textsc{FLORA}'s low-rank gradient moment estimation is unbiased, assuming projections are updated frequently enough; however, the application of random projection matrices yields an additive variance proportional to the number of trainable parameters. Thus, it is less suited for training large models from scratch. The authors address this caveat by setting a low update frequency, trading off bias for variance while focusing on small model training and fine-tuning tasks. Recently, \citet{shamshoum2024compAct} (\textsc{DropAct}) proposed applying a random projection (similarly to \textsc{FLORA}) to the linear layer's inputs during the forward phase before storing them for the backward phase to further reduce input memory for the weight gradient.
This approach reduces activation, gradients, and optimizer state memory as the gradients are only projected back for weight update.
We will later show (\cref{sec:finetune}) that \textsc{FLORA} typically leads to a higher loss than both \textsc{GaLoRE} and our method. We attribute this primarily to the high variance incurred by sampling high-dimensional random projections. 

In continuation to \textsc{GaLoRE}, \citet{liao2024galoreplus} (\textsc{GaLoRE}$+$)proposed reducing the SVD overhead by leveraging randomized SVD approximation for the top-$k$ singular vectors, along with sparse coding of the low-rank residual errors and sharing of projection matrices across attention heads weights in the same layer. The proposed method is focused on accelerating fine-tuning of LLMs, and its results are on par with \textsc{GaLoRE}. Finally, \citet{liang2024onlinedecent} suggested improving low-rank gradient estimators' performance by tracking the gradient subspace by an online PCA approach. This approach is designed to reduce the cost of frequent SVD, however, the online PCA cost is added to each iteration. Both methods are ultimately biased due to the accumulation of approximation error of the top-$k$ singular vectors.

\section{PLUMAGE: Min-Variance, Unbiased, and Low-Rank Estimator}
\label{sec:method}
The following section is dedicated to developing the min-variance unbiased low-rank gradient estimator, following similar design considerations as in \cite{alain2015variancereduction,wang2018atomo,chmiel2022minimum} while emphasizing favorable characteristics for optimization with low-rank gradients. Namely, a fixed target rank leading to a deterministic compute cost, and to use only a one-sided projection matrix to minimize required storage. Finally, we establish \textsc{PLUMAGE}'s efficient sampling strategy and develop the tools for deploying \textsc{PLUMAGE} in stateful optimizers such as \textsc{Adam}.

\subsection{MVUE: Min-Variance, Unbiased and Low-Rank Estimator}
\label{method}
Given the matrix $\mathbf{G} \in \mathbb{R}^{m\times n}$ and its singular values decomposition  $\mathbf{G}=\sum_{i=1}^{n}\sigma_{i}\mathbf{u}_{i}\mathbf{v}_{i}^{\top}$, where  $n\leq m$ and $\boldsymbol{\sigma}=\{\sigma_{i}\}_{i=1}^{n}$ is ordered from largest to smallest singular values. The classic low-rank estimator based on the top-$k$ singular vectors with $k\leq n$ is given by
\begin{equation}
\hat{\mathbf{G}}_{\mathrm{top-k}}= \sum_{i=1}^{k}\sigma_{i}\mathbf{u}_{i}\mathbf{v}^{\top}_{i}
\end{equation}
is known to have the minimal mean square error \cite{eckart1936approximation}. However, the error due to the truncation of the tail components results in a bias. It was also known that gradient bias is detrimental to the optimization process, as errors are accumulated over multiple training iterations \citep{chmiel2020neural,chmiel2022minimum}.
Therefore, we consider a low-rank gradient estimator with the general form
\begin{equation}
\label{basic-structure}
    \hat{\mathbf{G}}=\sum_{i=1}^{n}\frac{1}{p_{i}}I_{i}\sigma_{i}\mathbf{u}_{i}\mathbf{v}_{i}^{\top}
\end{equation}
%
where $I_{i}$ are indicator function for including the $i^\mathrm{th}$ singular vector while $p_{i}$ are scalar constants. We aim to find a distribution for $I_{i}$ such that the following properties hold:

    1) Unbiased gradient estimator:
        \begin{equation}%
            \mathbb{E}\hat{\mathbf{G}}=\mathbf{G}\Leftrightarrow p_{i}=\mathbb{E}I_{i}
        \end{equation}
    2) Deterministically of $k$-rank: 
        \begin{equation}
        \label{sparsity-req}
            \sum_{i=1}^{n}I_{i}=k \Rightarrow \sum_{i=1}^{n}p_{i}=k
        \end{equation}
    3) Minimal variance: 
        \begin{equation}
        \label{min-var-def}
            \min_{\hat{\mathbf{G}}} \mathbb{E}\mathrm{Tr}\left[\left(\hat{\mathbf{G}}-\mathbb{E}\hat{\mathbf{G}}\right)^{\top}\left(\hat{\mathbf{G}}-\mathbb{E}\hat{\mathbf{G}}\right)\right]
        \end{equation}
        %
%
To minimize the expected variance, we plug the estimator definition from \cref{basic-structure} into \cref{min-var-def} and, using the orthogonality of the singular vectors, we obtain $\mathbf{v}_{i}^{\top}\mathbf{u}_{j}=\delta_{ij}$,
\begin{align}
&\mathbb{E}\mathrm{Tr}\left[\sum_{i,j=1}^{n}\left(\frac{I_{i}}{p_{i}}-1\right)\left(\frac{I_{j}}{p_{j}}-1\right)\mathbf{u}_{i}\mathbf{v}_{i}^{\top}\mathbf{u}_{j}\mathbf{v}_{j}^{\top}\sigma_{i}\sigma_{j}\right]
=\mathbb{E}\mathrm{Tr}\left[\left(\sum_{i=1}^{n}\left(\frac{I_{i}}{p_{i}}-1\right)^{2}\sigma_{i}^2\mathbf{u}_{i}\mathbf{v}_{i}^{\top}\right)\right]
\\&=\mathbb{E}\left(\sum_{i=1}^{n}\left(\frac{I_{i}}{p_{i}}-1\right)^{2}\sigma_{i}^2\mathbf{v}_{i}^{\top}\mathbf{u}_{i}\right)=\sum_{i=1}^d\left[\mathbb{E}\left(\frac{I_{i}}{p_{i}}-1\right)^{2}\right]\sigma_{i}^{2}
=\sum_{i=1}^d\left[\left(\frac{1}{p_{i}}-1\right)\right]\sigma_{i}^{2}
\label{estimator-variance}
\end{align}

Taking into account the sparsity condition in \cref{sparsity-req}, we define $\boldsymbol{p}$ as the indicator probabilities vector that minimizes the variance in \cref{estimator-variance}. We find $\boldsymbol{p}$ by solving the following problem 
\begin{equation}
\min_{\boldsymbol{p}}\sum_{i=1}^{n}\left[\left(\frac{1}{p_{i}}-1\right)\right]\sigma_{i}^{2}\,\,\mathrm{s.t.}\,\,\text{\ensuremath{\sum_{i=1}^{n}p_{i}=k\,,p_{i}\in\left[0,1\right]}}
\end{equation}
which is equivalent to 
\begin{equation}
\label{optimization-problem}
\min_{\boldsymbol{p}}\sum_{i=1}^{n}\frac{\sigma_{i}^{2}} {p_{i}}\,\,\mathrm{s.t.}\,\,\text{\ensuremath{\sum_{i=1}^{n}p_{i}=k}},p_{i} \leq 1     
\end{equation}
Note that we cannot allow for $p_{i}=0$ since then the optimization objective is undefined.
To solve \cref{optimization-problem}, we write the Lagrangian 
\begin{equation}
\sum_{i=1}^{n}\frac{\sigma_{i}^{2}}{p_{i}}+\mu\sum_{i=1}^{n}p_{i}+\sum_{i=1}^{n}\lambda_{i}p_{i}
\end{equation}
where $\mu>0, \lambda_{i}>0$ are the Lagrangian factors, and after differentiating by $p_{i}$ we get 
\begin{equation}
    0=-\frac{\sigma_{i}^{2}}{p_{i}^{2}}+\mu+\lambda_{i} \Rightarrow p_{i}=\frac{\sigma_{i}}{\sqrt{\mu+\lambda_{i}}}
\end{equation}
Note that $\lambda_{i}>0$ only if we hit the inequality constraints, i.e., $p_{i}=1$. To satisfy the constraints, we first solve 
\begin{equation}
r^{*}\left(k,\boldsymbol{\sigma}\right)=\mathrm{arg}\min_{r\in\left\{ 0,..,d\right\} }r\,\,\mathrm{s.t.\,\,}\frac{\left(k-r\right)\sigma_{r+1}}{\sum_{i=r+1}^{n}\sigma_{i}}<1    
\end{equation}
Then, we return the following solution 
\begin{equation}
\label{eq:pi_equation}
    p_{i}=\begin{cases}
1 & ,\mathrm{if}\,\,i\leq r^{*}\\
\frac{\left(k-r^{*}\right)\sigma_{i}}{\sum_{j=r^{*}+1}^{n}\sigma_{j}} & ,\mathrm{if}\,\,i>r^{*}
\end{cases}\,.
\end{equation}
\begin{wrapfigure}{r}{0.4\textwidth}
\centering
\includegraphics[height=5cm,keepaspectratio, trim=1cm 1cm 1cm 1cm, clip]{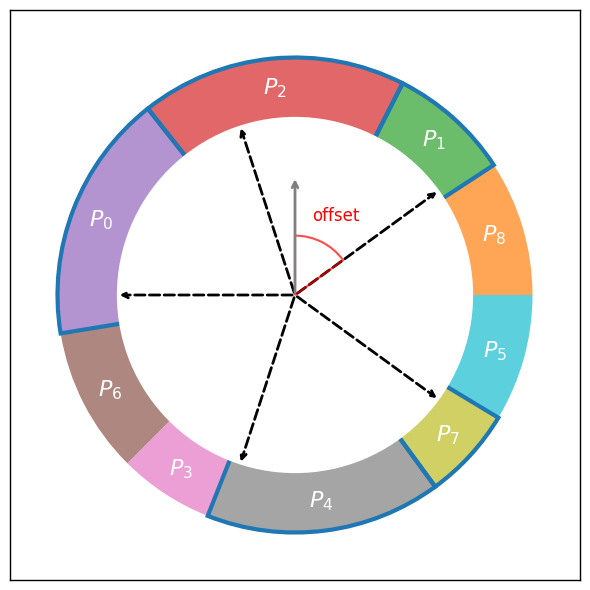}
\caption{``Wheel-of-Fortune" sampling with $k=5$ arms}
\label{fig:wheel_of_fort}
\end{wrapfigure}
It is important to note that $I_{i}$ need not be independent. Meaning, that we can sample the singular vectors however we want, just as long as we satisfy \cref{eq:pi_equation} (e.g. to satisfy the $k$-sparse condition in \cref{sparsity-req}). For example, we can first sample $i=1\ldots r^{*}$ (deterministically), then sample a $\left(k-r^{*}\right)$ sparse solution from the rest of the components. Specifically, let $k\leq n$ and $\{ p_i\}_{i=1}^{n}$ a series of the singular vectors inclusion probabilities, with $p_i\in(0,1)$ and $\sum^n_{i=1}p_i=k$. We wish to sample without replacement exactly $k$ distinct indices $I\triangleq \{i_1\ldots i_k \}$ such that $\forall i \in \{1\ldots n\}: \mathbb{P}(i\in I)=p_i$. To this end, we employ a ``wheel-of-fortune" sampling trick with $k$ equidistant arms \citep{wheeloffortune_mathoverflow} to efficiently sample projections. Namely, we shuffle the order of $p_i$, then each $p_i$ is represented by an angular sector that is proportional to $\frac{p_i}{\sum_{k=1}^n p_k}$. Finally, the arms are randomly rotated (with uniformly distributed shift) to extract all the indices in a single step. The sampling algorithm is illustrated in \cref{fig:wheel_of_fort}. Computing $\boldsymbol{p}$ and sampling can be implemented efficiently in $\mathcal{O}(\mathrm{min}(m,n))$ and is negligible compared to \textsc{SVD} computation.
%

\subsection{One-sided PLUMAGE}
\label{sec:plumage_projections}
Next, we move to construct an estimator using only a one-sided projection and show that our one-sided estimator maintains the MVUE properties. Without loss of generality, we assume our gradient estimator is based on a left-sided projection $\mathbf{P}$. The projection matrix is constructed by stacking the sampled singular values,
\begin{equation}
\label{eq:projection}
\mathbf{P} = \left[\,
\underbrace{\mathbf{u}_1 I_1}_{\text{column 1}} \quad
\dots \quad
\underbrace{\mathbf{u}_d I_d}_{\text{column } d}\,
\right] \,.
\end{equation}
Similarly, right-sided projections stack $\textbf{v}_i$ singular vectors as rows. Next,
given the diagonal scaling matrix $\mathbf{D}=\mathrm{Diag}(p_1,p_2,...,p_d)$, the one-sided estimator for a matrix $\mathbf{G}$ is given by
\begin{align}     \hat{\mathbf{G}}&=\mathbf{P}\mathbf{D}^{-1}\mathbf{P}^{\top}\mathbf{G} =\sum_{i=1}^{n}\frac{1}{p_{i}}I_{i}\mathbf{u}_{i}\mathbf{u}_{i}^{\top}\mathbf{G}\,.
\end{align}
Finally, we plug in $\mathbf{G}=\sum_{i=1}^{n}\sigma_{i}\mathbf{u}_{i}\mathbf{v}_{i}^{\top}\,.$ Through the orthonormality of the singular vectors, we recover 
\begin{align}
    \hat{\mathbf{G}}&=\sum_{j=1}^{n}\sigma_{j}\sum_{i=1}^{n}\frac{1}{p_{i}}I_{i}\mathbf{u}_{i}\mathbf{u}_{i}^{\top}\mathbf{u}_{j}\mathbf{v}_{j}^{\top}=\sum_{j=1}^{n}\sigma_{j}\sum_{i=1}^{n}\delta_{ij}\frac{1}{p_{i}}I_{i}\mathbf{u}_{i}\mathbf{v}_{j}^{\top}
    =\sum_{i=1}^{n}\frac{1}{p_{i}}I_{i}\sigma_{i}\mathbf{u}_{i}\mathbf{v}_{i}^{\top}\,.
    \label{one-side-is-two}
\end{align}
Since \cref{one-side-is-two} is equivalent to the original formulation from \cref{basic-structure}, the one-sided projection has the same properties as the two-sided variant.


\subsection{Incorporating \textsc{PLUMAGE} in Optimizers}
We apply \textsc{PLUMAGE} to compress the optimizer's first and second momentum, which are common statistics tracked by optimizers such as \textsc{Adam} and \textsc{SGDM}. 
\subsubsection{Low-rank moments and \textsc{Adam} weight update}
The low-rank moments are given by 
\begin{align}
\mathbf{M}^{\lfloor}_t&= \beta_1 \mathbf{M}^{\lfloor}_{t-1} + \left(1-\beta_1\right) \mathbf{P}^\top_t\mathbf{G}_t \\
\mathbf{V}^{\lfloor}_t&= \beta_2 \mathbf{V}^{\lfloor}_{t-1} + \left(1-\beta_2\right) \left(\mathbf{P}^\top_t\mathbf{G}_t\right)^{\circ 2} ,
\end{align}
and the simplified low-rank \textsc{Adam} update rule is
\begin{equation}
    {\mathbf{W}}_{t+1} = {\mathbf{W}}_t - \eta \mathbf{P}_t\mathbf{D}_t^{-1} \frac{{\mathbf{M}^{\lfloor}}_t}{\sqrt{{\mathbf{V}^{\lfloor}}_t} + \epsilon} \,,
\end{equation}
where $\mathbf{P}_t \in \mathbb{R}^{m \times r}$ denotes the left projection matrix composed from the singular vectors of $\mathbf{G}_t$ and $\mathbf{D}_t$ is similarly the diagonal matrix containing its sampled singular values, as explained in \cref{sec:plumage_projections}.
\subsubsection{Design considurations}
We wish to amortize costly \textsc{SVD} by reusing the same projection over multiple optimizer steps, similarly to \citet{galore}. Given the hyperparameter $\tau\in \mathbb{N}^+$, the interval length between subsequent \textsc{SVD}. For notation convenience, we define the index mapping $k_t=\lfloor\nicefrac{t}{\tau}\rfloor$. We define the optimizer statistics as  
\begin{align}
\mathbf{M}^{\lfloor k_t}_t&= \beta_1 \mathbf{M}^{\lfloor k_t}_t + \left(1-\beta_1\right) \mathbf{P}_{k_t}\mathbf{G}_t \\
\mathbf{V}^{\lfloor k_t}_t&= \beta_2 \mathbf{V}^{\lfloor k_t}_t + \left(1-\beta_2\right) \left(\mathbf{P}_{k_t} \mathbf{G}_t\right)^{\circ 2}\\
{\mathbf{W}}_{t+1} &= {\mathbf{W}}_t - \eta \mathbf{P}_{k_t}\mathbf{D}_{k_t}^{-1} \frac{{\mathbf{M}^{\lfloor k_t}_t}}{\sqrt{{\mathbf{V}^{\lfloor k_t}_t}} + \epsilon} \,.
\label{plumage_base}
\end{align}
The projection $\mathbf{P}_{k_t}$ and scaling factors $\mathbf{D}_{k_t}$ are sampled once after computing \textsc{SVD} over $\mathbf{G}_{k_t}$ according to the probability $\boldsymbol{p}_{k_t}$ computed according to \cref{eq:pi_equation}. 
In this case, $\boldsymbol{p}_{k_t}$ is no longer proportional to $\mathbf{G}_{k_t}$'s singular values. Thus, we lose the minimum variance promise of \textsc{PLUMAGE}; however, the estimate remains unbiased so long as $\mathrm{\#total\_training\_steps}\gg\tau$ (i.e., the projections are resampled often enough, albeit additional SVD is not required as we explain next). Moreover, note that \textsc{PLUMAGE} needs only to store the diagonal elements in $\mathbf{D}_{k_t}$ up to the chosen rank, contributing to minor memory overhead compared to the baseline \textsc{GaLoRE} memory cost. 

If the gradient subspace changes slowly during optimization, we can still benefit from variance reduction. In addition, one can also define the $\kappa\in \mathbb{N}^+$ hyperparameter to introduce a resampling interval that refreshes the projection and scaling coefficients without computing the \textsc{SVD} (i.e., $\kappa\leq\tau$). This can help balance subspace exploration and \textsc{SVD} compute overhead under the uncertainty of how well the sampled subspace retains the information from the observed gradients. In our experiments, sampling projection once throughout the $\mathrm{SVD}$ update interval works well. Yet, we found that overly frequent projection updates may hurt the utility of the moment statistics in the optimization process. We leave the exploration of this tradeoff for future work and set $\kappa=\tau$ in our experiments. In addition, we explored methods for measuring the subspace correlation with the recent gradients in the appendix.
\subsubsection{Statistics realignment of the first and second moments}
\label{sec:realignment}
Stateful optimizers such as \textsc{Adam}, when using low-rank estimators such as \textsc{GaloRE}, suffer from a crucial subspace alignment issue when the gradient's projection is updated during training. In essence, the low-rank statistics represent moment estimates tracked in different subspaces. Formally, let $\mathbf{G}^{\lfloor t}=\mathbf{P}^\top_t\mathbf{G}$
then, $\mathbf{M}^{\lfloor t}=\mathbb{E}\left[\mathbf{G}^{\lfloor t}\right]$ and $\mathbf{V}^{\lfloor t}=\mathbb{E}\left[\left(\mathbf{G}^{\lfloor t} \right)^{\circ 2} \right]$. However, given two arbitrary projection metrices $\mathbf{P}_1\in \mathbb{R}^{m\times r_1}$ and $\mathbf{P}_2\in \mathbb{R}^{m\times r_2}$, it is clear that $\mathbf{M}^{\lfloor 1} \neq \mathbf{M}^{\lfloor 2}$ and $\mathbf{V}^{\lfloor 1} \neq \mathbf{V}^{\lfloor 2}$
For instance, if $\mathbf{P}_1\perp\mathbf{P}_2$, then $\mathbf{M}^{\lfloor 1}$ has no information on $\mathbf{M}^{\lfloor 2}$.  
Moreover, misalignment can occur even if the subsequent projections span the same subspace (e.g., a rotation of the singular vectors). Therefore, updating the projections without statistics realignment may lead to spurious weight updates in the following weight updates (until the new data is sufficiently represented in the first and second moment estimators).
Similarly to \citet{hao2024flora}, one can initialize $\mathbf{M}^{\lfloor 2}$, after updating the projection, by projecting the previous estimate onto the shared subspace between the two projections to realign the first-moment statistics, 
\begin{equation}
\label{eq:MP}
    {\mathbf{M}}^{\lfloor 2} \approx \mathbf{P}_2^\top \mathbf{P}_1\mathbf{M}^{\lfloor 1}
\end{equation}
Notably, unlike \citet{hao2024flora}, under \textsc{PLUMAGE}'s $\mathbf{P}_i$ are orthonormal matrices. Thus, $\mathbf{B}=\mathbf{P}_2^\top \mathbf{P}_1$ projects $\mathbf{M}^{\lfloor 1}$ onto the intersection of subspaces induced by the projections $\mathbf{P}_i$. The resulting ${\mathbf{M}}^{\lfloor 2}$ magnitude will be proportional to the cosine of the principal angle between the two subspaces since $\Vert\mathbf{B}\Vert\leq1$. Moreover, when $\mathbf{P}_i$ spans the same subspace, the transformation rotates the moment without distortion since $\Vert\mathbf{B}\Vert=1$.
For $\mathbf{V}^{\lfloor 2}$, one cannot simply initialize $\mathbf{V}^{\lfloor 2}=\mathbf{B}\mathbf{V}^{\lfloor 1}$ after the change in projections, since it will produce negative entries. Since finding the optimal inverse projection is not trivial, we devise an approximation for $\mathbf{V}^{\lfloor 2}$ as an initial guess.
\begin{align}
{\mathbf{V}}^{\lfloor 2}_{ij}=\mathbb{E}\left[\mathbf{B}\mathbf{G}^{\lfloor 1}\right]^{2}_{ij}
=\sum_{k,l}B_{ik}B_{il}\mathbb{E}\left[G^{\lfloor 1}_{kj}G^{\lfloor 1}_{lj}\right]
\approx\sum_{l, k}B_{ik}B_{il}\mathbb{E}\left[\left(G^{\lfloor 1}_{kj}\right)^2\right] 
=\Biggr[\mathbf{B}^{\circ 2}\mathbf{V}^{\lfloor 1}\Biggr]_{ij}\,,
\label{v''_approx_A}
\end{align}
where in the $\approx$ step we approximated the gradient's second-moment matrix to be diagonal, i.e., $\mathbb{E}\left[G_{kj}G_{lj}\right]\approx0$ for $k \neq l$. Such a diagonal approximation is commonly used in both theory and practice \citep{kingma2017adammethodstochasticoptimization,amari2018fisherinformationnaturalgradient}.
In \cref{sec:ablation_study}, we empirically demonstrate that our initialization produces better results than simply assuming $\mathbf{M}^{\lfloor 1}\approx \mathbf{M}^{\lfloor 2}$ and $\mathbf{V}^{\lfloor 1}\approx \mathbf{V}^{\lfloor 2}$ when updating the gradient projection.


A full algorithm demonstrating how to incorporate \textsc{PLUMAGE} in \textsc{Adam} optimizer is given in \cref{alg:GaLoRE-v2} in the appendix.

\section{Experiments}
\subsection{PLUMAGE Ablation Study} \label{sec:ablation_study}
Similarly to \citet{galore}, we evaluate the optimization performance of our Adam-based \textsc{PLUMAGE} optimizer by pre-training LLaMA with varying sizes on the C4 English dataset 
from scratch. In all C4 English experiments, we rely on the processed version of the ``Colossal Cleaned Crawl", which is readily available in the Huggingface datasets library \citep{wolf2019huggingfaces,allenai-c4,raffel2019exploring-commoncrawl}. As in \citet{galore}, we split the full-length training and validation examples into segments with 256 tokens per example using the T5 tokenizer from \citet{raffel2019exploring-commoncrawl}.

\begin{wraptable}{r}{0.5\textwidth}
\caption{Ablation: terminal loss and accuracy observed on Llama 130M after 2.6B Tokens.}
\label{tab:ablation}
\centering
\resizebox{\linewidth}{!}{%
\begin{tabular}{lccc}
\toprule
  & Train Loss & Eval Loss & Accuracy \\
\midrule
 \textsc{Adam} & 3.256 & 3.256 & 38.7 \\
 \textsc{GaLoRE} & 3.473 & 3.471 & 36.5 \\
 \hline
 \textsc{PLUMAGE} & 3.402 & 3.400 & 37.1 \\
 $\mathrm{PLUMAGE_{MP}}$ & 3.386 & 3.385 & 37.3 \\
 $\mathrm{PLUMAGE_{S/MP}}$ & 3.378 & 3.377 & 37.4 \\
\bottomrule
\end{tabular}%
}
\end{wraptable}
Specifically, we set the learning rate to $10^{-3}$  and $\beta_1,\beta_2\leftarrow0.9,0.999$, while the low-rank gradient estimators' additional hyperparameters are set to $r,\tau,\alpha\leftarrow128,200,1.0$, where $\tau$ is the interval between projection updates, and $\alpha$ is the learning rate scale applied only to the low-rank layer weights in \textsc{GaLoRE}

The ablation results (\cref{tab:ablation},\cref{fig:ablation_llama130m}) include the base \textsc{PLUMAGE} (\cref{plumage_base}), the 'first \textbf{M}omentum \textbf{P}rojection realignment' method (\cref{eq:MP}) tagged with $\mathrm{MP}$, and the`\textbf{S}econd moment realignment' (\cref{v''_approx_A}) method tagged with $\mathrm{S/MP}$.


An optimization gap between the low-rank gradient optimizers and the full-rank \textsc{Adam} is apparent.
However, $\mathrm{PLUMAGE_{S/MP}}$ shrinks the gap between the \textsc{GaLoRE} and Adam baseline by a factor of $\sim2$ when not tuning the hyperparameters from the baseline training regime (that \citet{galore} found to be optimal for \textsc{Adam}).
\subsection{PLUMAGE debias impact}
\begin{figure}
    \begin{subfigure}[t]{0.48\textwidth}
    \centering
    \includegraphics[height=5cm,keepaspectratio]{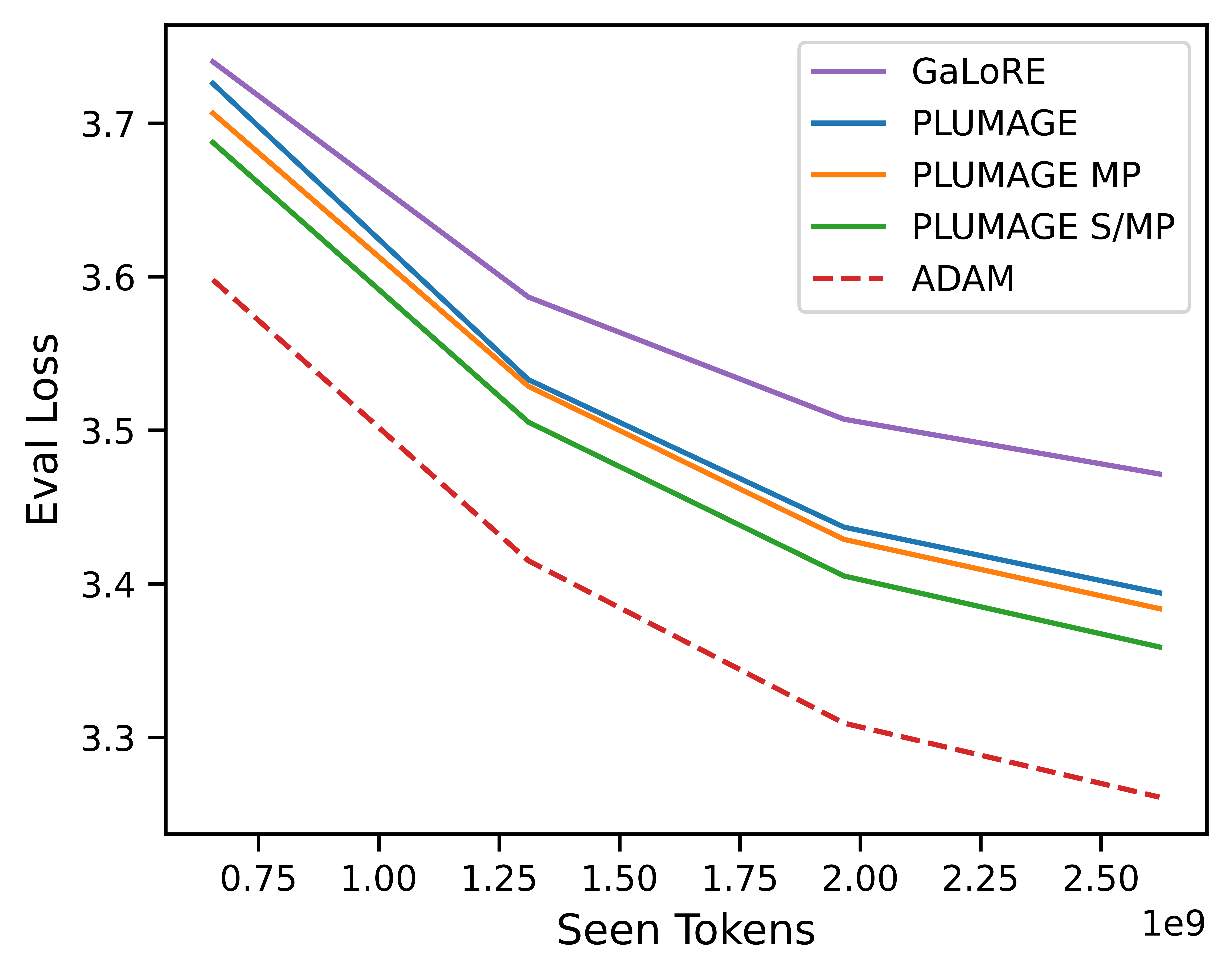}
    \caption{Llama 130M Ablation: perplexity vs. tokens.}
    \label{fig:ablation_llama130m}
    \end{subfigure}
    \hfill
    \begin{subfigure}[t]{0.48\textwidth}
    \centering
    \includegraphics[height=5cm,keepaspectratio]{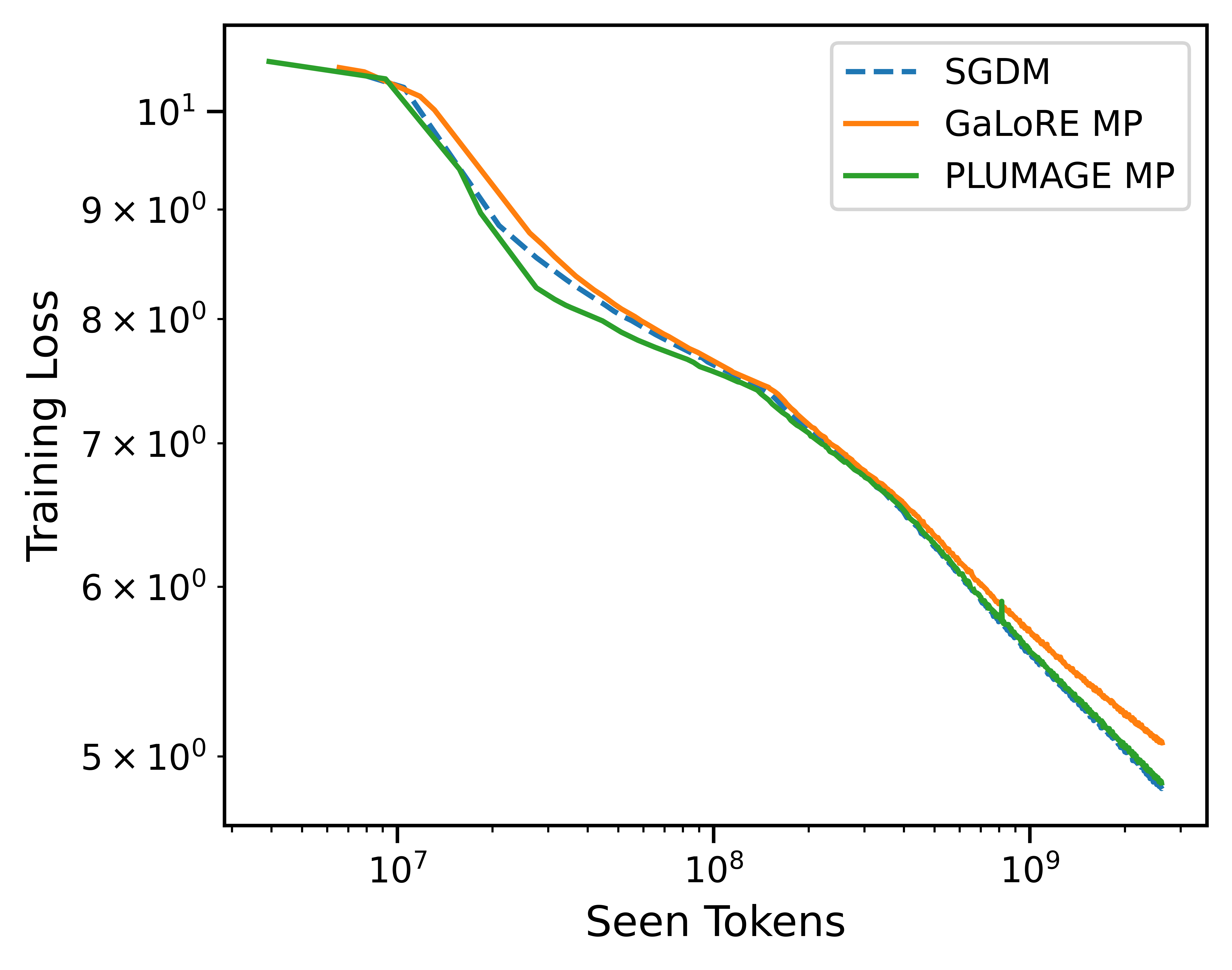}
    \label{fig:llama350_sgdm}
    \caption{Llama 350M C4 loss, w/o second moment.}
    \end{subfigure}
    %
    \caption{}
\end{figure}

While \textbf{PLUMAGE} is unbiased, applying it to \textsc{Adam} does not produce unbiased gradients due to the use of the non-linear adaptive scaling factor based on the $\mathbf{V}$ statistic. The $\mathbf{V}$ term contributes to gradient bias with the \textsc{Adam} update as well, thus degradation observed in \cref{tab:ablation} is expected. 
To better understand how bias impacts optimization with low-rank gradients, we train a medium-sized Llama variant with 350M parameters on 2.6B tokens from the C4 English dataset as before. However, since the adaptive step size is biased, we use stochastic gradient descent with momentum ($\mathrm{SGDM}$). This is achieved by disabling the second-moment term in Adam and comparing it with the adapted versions of $\mathrm{PLUMAGE_{MP}}$ and $\mathrm{GaLoRE_{MP}}$ variants. We use momentum $\beta=0.9$ for all optimizers and a learning rate of $0.1$. In addition, we fix the projection intervals to 200 steps and the rank to 128, while the hidden dimension of the model is 1024. In \cref{fig:llama350_sgdm}, we observe the training loss on a log scale. Our version of $\mathrm{SGDM + PLUMAGE_{MP}}$ convergence rate is on par with the full-rank SGDM optimizer, while it is clear that $\mathrm{SGDM + GaLoRE_{MP}}$ bias is impeding the optimization.

%

\subsection{Fine-tuning with Low-Rank Gradients}
\label{sec:finetune}
Following \citet{galore} and \citet{hao2024flora}, we evaluate the low-rank optimization on the standard GLUE natural understanding benchmark \citep{wang2019gluemultitaskbenchmarkanalysis}. 
We fine-tune all parametrized layers in the Roberta base model \citep{liu2019robertarobustlyoptimizedbert} except for its embedding layers, while low-rank gradient estimators are used for all linear layers' weights.
We fix the rank $r = 8$, and reuse the hyperparameters as suggested in \citet{galore}. In particular, the update interval $\tau = 100$ and the recommended $\alpha$ scales for \textsc{GaLoRE} (recall that \textsc{PLUMAGE} does not use $\alpha$). We optimize the model for $30$ epochs for each task with individually tuned learning rates for each method as done in \citet{galore}. Finetuning longer than the usual number of epochs ensures the loss has converged. This allows us to compare the terminal loss between the different estimators to evaluate the estimators' impact on the optimization gap. The results are reported in \cref{tab:roberta-base}, where for each task we show the best metrics for accuracy and loss. Again, a training error gap exists in all low-rank fine-tuning methods since a bias exists in all adaptive optimizers. It is also notable that replacing \textsc{GaLoRE} with \textsc{PLUMAGE} leads to lower or equivalent terminal loss and next token prediction accuracy.
\begin{table*}[!ht]
\caption{Fine-tuning RoBERTa-Base on GLUE benchmark. We present the best terminal accuracy (and loss) metrics for each task after 30 epochs over 5 learning rates for each optimizer. }
\label{tab:roberta-base}
\centering
\resizebox{\textwidth}{!}{%
\begin{tabular}{lcccccccc|c}
\toprule
 & MNLI & QQP & QNLI & SST-2 & CoLA & STS-B & MRPC & RTE & Mean \\
\midrule
ADAM & 87.60 (0.004) & 89.35 (0.002) & 92.86 (0.002) & 94.27 (0.004) & 63.58 (0.004) & 90.92 (0.03) & 92.93 (0.001) & 75.81 (0.003) & 85.91 (0.004) \\
\midrule
$\mathrm{FLORA}$ &87.07 (0.25) & 87.77 (0.15) & 92.33 (0.15) & 94.84 (0.07) & 59.31 (0.13) & 89.95 (0.22) & 91.42 (0.16) & 71.84 (0.24) & 84.32 (0.16) \\
\textsc{GaLoRE} & 85.78 (0.29) & 85.46 (0.21) & 91.95 (0.11) & 92.78 (0.08) & 62.32 (0.07) & 90.75 (0.09) & 91.36 (0.02) & 78.70 (0.04) & 84.89 (0.11) \\
\midrule
$\mathrm{PLUMAGE_{S/MP}}$ & 
87.58 (0.22) & 88.23 (0.13) & 92.62 (0.06) & 94.72 (0.05) & 61.57 (0.03) & 90.94 (0.09) & 91.29 (0.005) & 78.34 (0.02) & 85.66 (0.08) \\
\bottomrule
\end{tabular}%
}
\end{table*}

\subsection{Pre-training language models from scratch}
The pre-training process of LLMs involves digesting trillions of tokens, setting a high bar for model quality. It is challenging to replicate such experiments, particularly on consumer-grade equipment. However, exploring the dynamics of training large models from scratch is crucial for facilitating research into more efficient methods for training at scale. Thus, we attempt to train the models to a point where we can gain insight into the performance of different low-rank gradient optimizers. 

We follow \citet{galore} and pre-train several Llama variants with 130M,350M, and 1B parameters on the C4 English dataset \citep{allenai-c4,raffel2019exploring-commoncrawl}. We reuse the training regime and the best hyperparameters reported in \cite{galore} for \textsc{GaLoRE} and \textsc{Adam}. All optimizers use the default $\beta_1,\beta_2,\epsilon\leftarrow0.9,0.999,1e-8$. For \textsc{PLUMAGE} we reuse \textsc{Adam}'s learning rate. In addition, we use a cosine decay schedule with a terminal value of $10\%$ from the maximum learning rate and a linear warmup period for 10\% of the total number of steps. 
\begin{wraptable}{r}{0.5\textwidth}
    \caption{Token-level validation perplexity for pre-training with low-rank gradient estimators.}
    \label{tab:pretraining_c4}
    \resizebox{\linewidth}{!}{%
    \begin{tabular}{c|ccc}
        \toprule
        \textbf{Method} & \textbf{130M} & \textbf{350M} & \textbf{1B} \\
        \hline
        \textsc{Adam} & 25.95 & 19.02 & 14.3 \\
        \textsc{GaLoRE}     & 30.18   & 24.08 & 17.03 \\
        $\mathrm{PLUMAGE_{S/MP}}$& 28.73 & 21.81 & 16.29 \\
        \hline
        Rank/Hidden &  256/768 & 256/1024 & 512/2048 \\
        Tokens Seen & 2.6B & 7.9B & 26.2B \\
        Optimizer Steps & 20K & 60K & 100K \\
        Warmup Steps & 2K & 6K & 10K \\ 
        \bottomrule
    \end{tabular}
    }
\end{wraptable}

We run a learning rate grid search for the 130M and 350M models over $0.05,0.01,0,005,0.0025,0.001,0.0075$ to confirm the choice of the learning rate for \textsc{PLUMAGE}. We find that $0.001$ consistently produces close to the best result, similar to the vanilla \textsc{Adam}. 
To match the computational and memory budget between the methods, both low-rank gradient estimators use the same fixed projection update interval of $\tau=200$ steps. We also match the rank for each model according to \citet{galore}. 
Our approach, as shown in \cref{tab:pretraining_c4}, consistently outperforms \textsc{GaLoRE} --- without having to sweep over additional hyperparameters and without impacting the total train time. Moreover, training PLUMAGE was stable over multiple seeds, while in our experiments \textsc{GaLoRE} suffers from sharp fluctuations in the training loss, as well as training failures as the training loss diverged in some seeds.

\section{Discussion}

\textbf{Summary.} In this work, we present \textsc{PLUMAGE}, a method to improve \textsc{GaLoRE} with similar computational and memory costs. Our method samples projections without replacement and yields a $k$-sparse and unbiased minimum variance estimator of the gradient. Our approach improves the accumulated projected statistics that are updated periodically during training. We demonstrate that our method can be used as a drop-in replacement for the full training procedure (without needing to re-tune the learning rate). During fine-tuning, we obtain similar accuracy as with standard (full-rank) training. During pre-training, we significantly decrease the degradation of \textsc{GaLoRE} compared to full-rank training. Notably, \textsc{PLUMAGE} has the potential to improve storage and communication in a wider context. The following applications do not dramatically change the optimization process from our experimental setup, yet we leave their in-depth exploration to future work.


\textbf{Activation memory compression.} 
Similarly to \citet{shamshoum2024compAct}, applying strictly right-sided \textsc{PLUMAGE} projections directly to linear layers' inputs during the forward phase can lead to low-rank storage of activation, gradient, and optimizer state. This is mathematically equivalent to \cref{plumage_base}, albeit gradient projection cost is added to each training forward pass instead of just once per optimizer step, and gradient SVD cannot be done with gradient accumulation without additional memory. 

\textbf{Low-Rank Gradient Accumulation Buffers.} 
The gradient size impacts both memory and communication costs throughout training.
Moreover, when training with mixed precision, gradients are typically kept at higher precision. Thus, when both strategies are used for LLM training, gradient buffers are often distributed across the collective devices' memory at the cost of significant communication overhead \citep{Rajbhandari_2020deepspeed, Zhao_2023fsdp}.
Similarly to \cite{hao2024flora, Grass-sparse-grad2024}, low-rank gradients via \textsc{PLUMAGE} can be accumulated directly into low-rank buffers as long as the projection is fixed between weight updates. Moreover, similarly to \citet{wang2018atomo}, \textsc{PLUMAGE} offers the additional benefit of using a device-specific low-rank projection to recover an estimate of the global batch gradient in a data-parallel setting, trading off compute overhead with communication and memory costs while halving the size of the communication overhead of \textsc{ATOMO} (since only one-sided projection is used).


\bibliography{main}
\bibliographystyle{abbrvnat}

\ifshowappendix
\newpage
\appendix
\section*{Technical Appendices and Supplementary Material}
\section{Measuring projection fitness via principal angles and adaptive projection intervals}
Since applying $\mathrm{SVD}$ to each gradient at every step is costly, similarly to \textsc{GaLoRE}, we amortize the cost of $\mathrm{SVD}$ over multiple optimization steps, reusing the projection from an old gradient and updating it within some interval. However, since projected state optimizers \citep{galore,liang2024onlinedecent} converge despite using uncorrected statistics with alternating projections, we hypothesize that the dominant directions are relatively stable at least in some layers. \citet{galore} also suggested that some layers require less frequent updates than others and proposed a per-layer adaptive interval controller to reduce the computational overhead of \textsc{SVD}. Specifically, the controller computes the cosine similarity between the first singular vectors of two subsequent projections to determine if the interval can be extended based on a fixed threshold. We argue that the suggested metric can be sensitive and unreliable. For example, if the ordering of singular vectors changes while the subspace remains unchanged, the proposed metric will return $0$. In addition, when the correlation between subsequent subspaces breaks, the metric cannot be used to shrink the interval to reflect uncertainty.

We devise an alternative approach based on the framework of principal angles \cite{Zhu2012PrincipalAB}. 
Namely, we calculate the cosine of the principal angle between the spanning subspaces can be retrieved by computing the singular values of $\mathbf{P}^\top_1\mathbf{P}_2 \in \mathbb{R}^{r_1\times r_2}$ as follows,
\begin{equation}
    \boldsymbol{\sigma}\leftarrow\mathrm{SVD}\left(\mathbf{P}^\top_1\mathbf{P}_2 \right).
        \label{eq:adaptive_p_angle_1}
\end{equation}
Finally, we take the mean cosine angle to represent the intersection between the two induced subspaces.
\begin{equation}
%
    \rho = \mathrm{Mean}(\boldsymbol{\sigma})
    \label{eq:adaptive_p_angle}
\end{equation}

\begin{wraptable}{r}{0.5\textwidth}
    \caption{Token-level validation perplexity for pre-training with low-rank gradient estimators.}
    \label{tab:pretraining_c4_with_adaptive}
    \resizebox{\linewidth}{!}{%
    \begin{tabular}{c|ccc}
        \toprule
        \textbf{Method} & \textbf{130M} & \textbf{350M} & \textbf{1B} \\
        \hline
        \textsc{Adam} & 25.95 & 19.02 & 14.3 \\
        \textsc{GaLoRE}     & 30.18   & 24.08 & 17.03 \\
        $\mathrm{PLUMAGE_{S/MP}}$& 28.73 & 21.81 & 16.29 \\
        \hline
        $\mathrm{PLUMAGE_{S/MP/A}}$ & 29.26 & 21.79 & 16.24  \\
        \hline
        Rank/Hidden &  256/768 & 256/1024 & 512/2048 \\
        Tokens Seen & 2.6B & 7.9B & 26.2B \\
        Optimizer Steps & 20K & 60K & 100K \\
        Warmup Steps & 2K & 6K & 10K \\ 
        \bottomrule
    \end{tabular}
    }
\end{wraptable}
This approach offers a principled and robust measure for the overlap between subsequent projections during training. This approach allows our controller to manage the period between projection updates reliably. One specific advantage is the ability to reduce the interval when the assumptions regarding subsequent overlap break during training.
In practice, we define a simple hysteresis controller with 3 threshold hyperparameters, $\gamma_\mathrm{shrink},\gamma_\mathrm{expand},\gamma_\mathrm{reset}$ that can be set by observing how well the training loss with a low-rank gradient estimator follows the loss of a standard optimizer and the recoded values of the principle angles with some fixed interval. The adaptive controller is given in \cref{alg:controller}. In practice, the overlap can be approximated well with only the top 64 singular vectors from the old sampled projection matrix and the top 64 singular vectors from the freshly computed singular vectors of the new gradient (without sampling) to reduce the computational overhead of \textsc{SVD} when extracting the mean cosine principal angles. In addition, we find that allowing the interval to grow unconstrained undermines the original purpose of accelerating training, as it leads to degradation in the loss convergence rate and terminal value. Thus, we set a $\tau_{max}=5\%$ of the total steps while $\tau_\mathrm{min}=\tau_{\mathrm{initial}}=200$. Finally, we set $\gamma_\mathrm{shrink},\gamma_\mathrm{expand},\gamma_\mathrm{reset}\leftarrow 0.4,0.6,0.3$ by observing the statistics during training, as can be seen in \cref{fig:mean_cosine_angle_per_layer,fig:alpha_and_r_per_layer}. 
We report the results in \cref{tab:pretraining_c4_with_adaptive}. Using our conservative adaptive interval hyperparameters, the equal-weighted average interval length over all layers in the models grew from the initial interval of 200 steps to $\sim425$ steps and $\sim1000$ steps for the 1B and 350M models. We observed minor terminal loss improvements in 1B and 350M models, potentially due to improved moment estimation with longer intervals. Ultimately, this avenue requires significant manual tuning to produce real train time gains while avoiding degradation in loss. Thus, we leave further exploration of this topic for future study utilizing larger models where the \textsc{SVD} overhead dominates the computation time.

\begin{algorithm}[tb]
    \caption{Adaptive projection interval}
   \label{alg:controller}
\begin{algorithmic}
   \STATE {\bfseries inputs:} subsequent projections $\textbf{P}_{t-1},\textbf{P}_{t} \in \mathbb{R}^{m \times r}$, interval configurations $\tau_\mathrm{min},\tau_\mathrm{max},\tau_\mathrm{initial}$, and thresholds $\gamma_\mathrm{reset}\leq \gamma_\mathrm{shrink}<\gamma_\mathrm{expand} \in [0,1]$
   \STATE {\bfseries initialize:} $\tau_0 \leftarrow \tau_\mathrm{initial}$,$t \leftarrow 0$ 
   \STATE $\rho_t \leftarrow \mathrm{mean\_cosine\_principle\_angle}(\textbf{P}_{t-1},\textbf{P}_{t})$ \RTCOMMENT{\cref{eq:adaptive_p_angle_1,eq:adaptive_p_angle}}
   \IF{$\rho_t < \gamma_\mathrm{reset}$}
        \STATE $\tau_t\leftarrow \tau_\mathrm{min}$
   \ELSIF{$\rho_t < \gamma_\mathrm{shrink}$}
        \STATE $\tau_t\leftarrow \mathrm{max}(\tau_{t-1}/2,\tau_\mathrm{min})$
    \ELSIF{$\rho_t > \gamma_\mathrm{expand}$}
        \STATE $\tau_t\leftarrow \mathrm{min}(2\cdot\tau_{t-1},\tau_\mathrm{max})$
   \ELSE
   \STATE $\tau_t\leftarrow \tau_{t-1}$
   \ENDIF
   \STATE $t\leftarrow t+1$
   \STATE {\bfseries return:}$\tau_t$
   
\end{algorithmic}
\end{algorithm}
\section{Additional qualitative experiments}
\begin{figure}
\begin{subfigure}[t]{0.48\textwidth}
    \includegraphics[height=5cm,keepaspectratio]{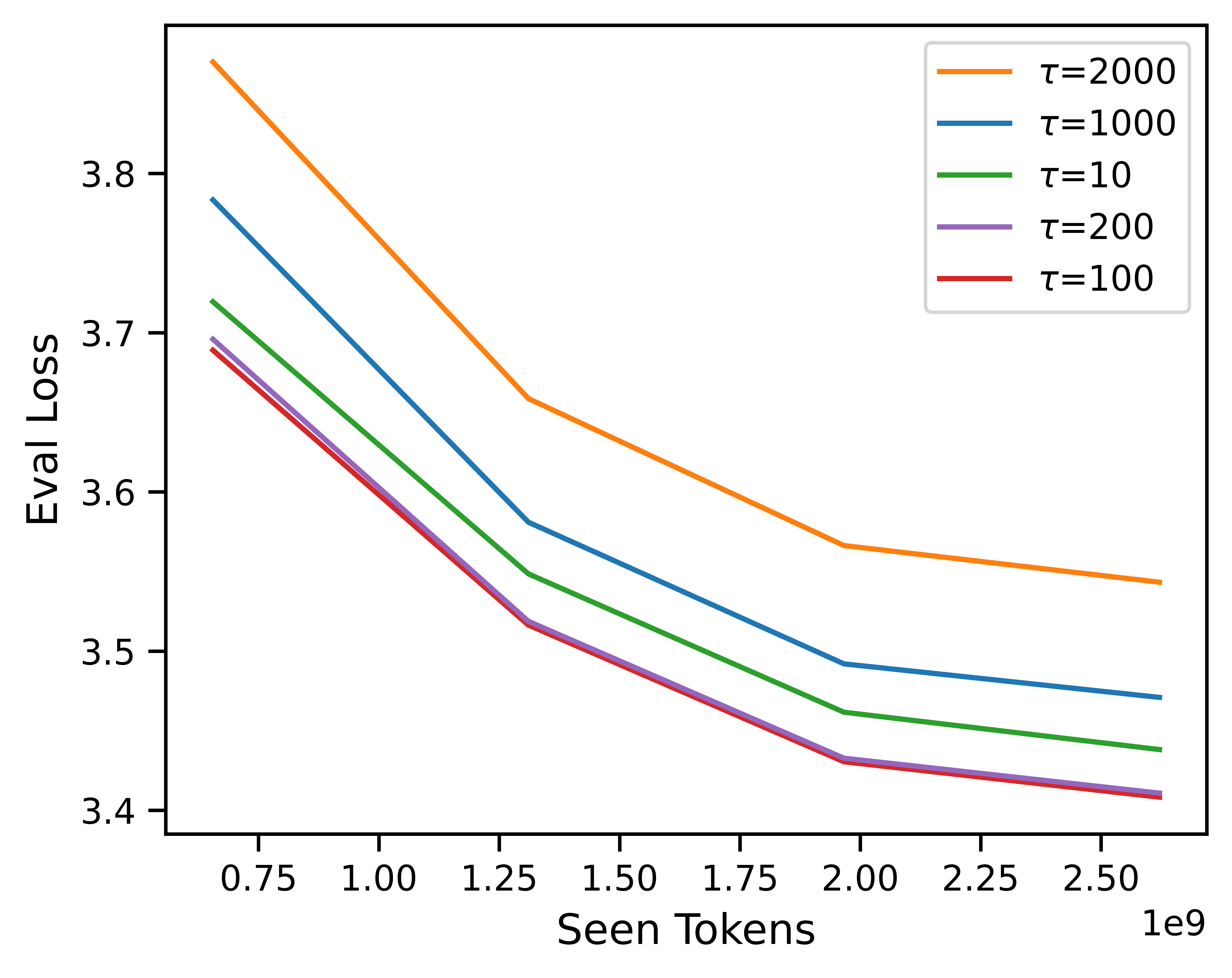}
    \caption{Impact of $\tau$ (\textsc{SVD} interval)}
    \label{fig:gap_ablation}
\end{subfigure}
\hfill
\begin{subfigure}[t]{0.48\textwidth}
\includegraphics[height=5cm,keepaspectratio]{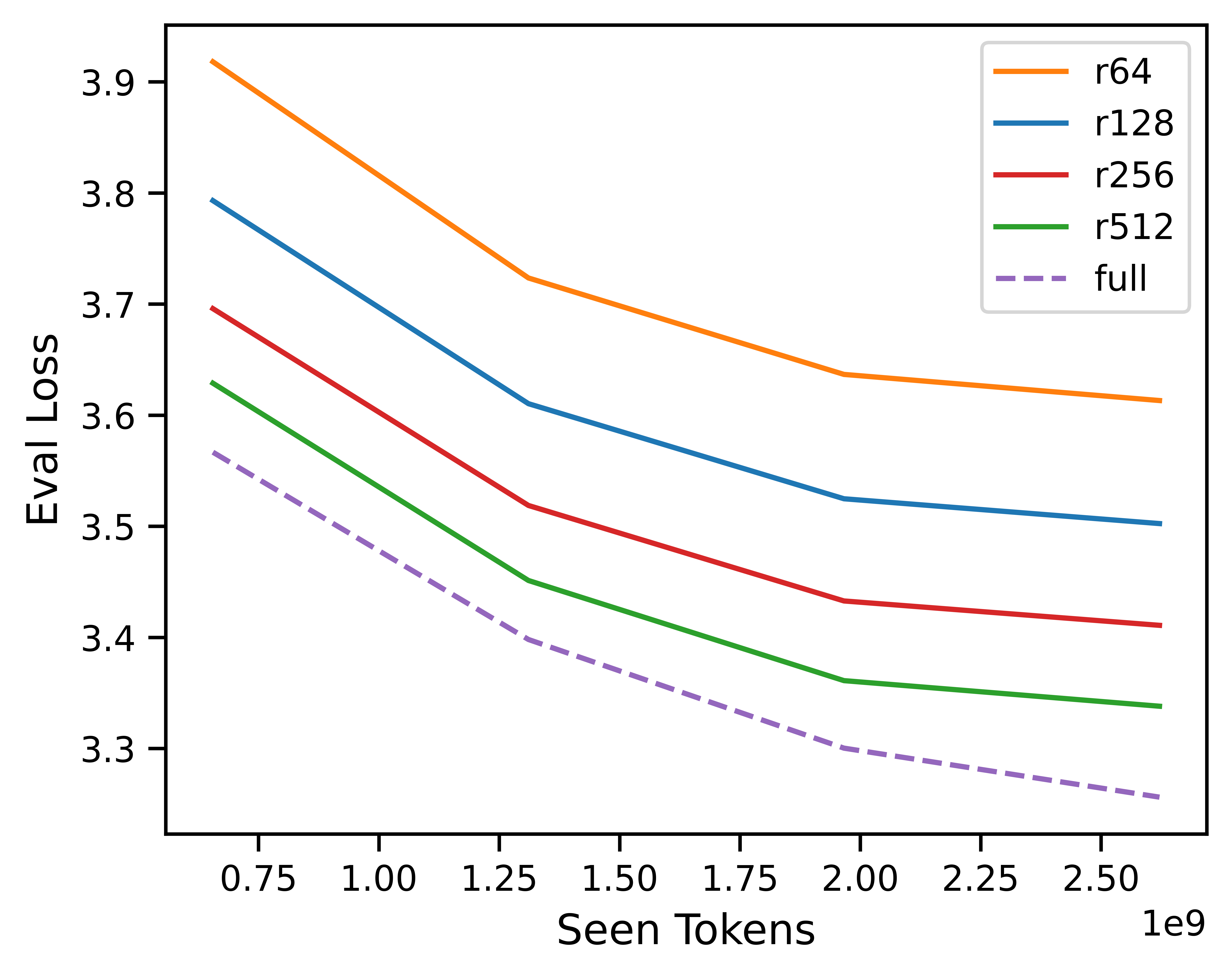}
    \caption{Impact of rank budget}
    \label{fig:rank_ablation}
\end{subfigure}
\caption{Ablation studies comparing validation loss Llama 130m on C4}
\label{fig:aux_ablation}
\end{figure}
\subsection{Rank and interval ablation studies}
We perform additional ablation experiments on the rank and \textsc{SVD} interval impact on the Llama 130M variant. The results are presented in \cref{fig:aux_ablation}. Similarly to \citet{galore}, we find that setting the gap interval too short or too wide leads to degradation in the optimization process. The short intervals potentially lead to poor first and second-moment estimates due to frequent projection updates. In addition, since $\mathbf{B}=\mathbf{P}_2^\top \mathbf{P}_1\leq 1$ (see \cref{sec:realignment}), the effective learning rate is also reduced. Finally, the choice of rank should be as large as one can fit onto the device memory, as can be seen in \cref{fig:rank_ablation}. 

\begin{figure}
    \centering
    \includegraphics[width=\linewidth]{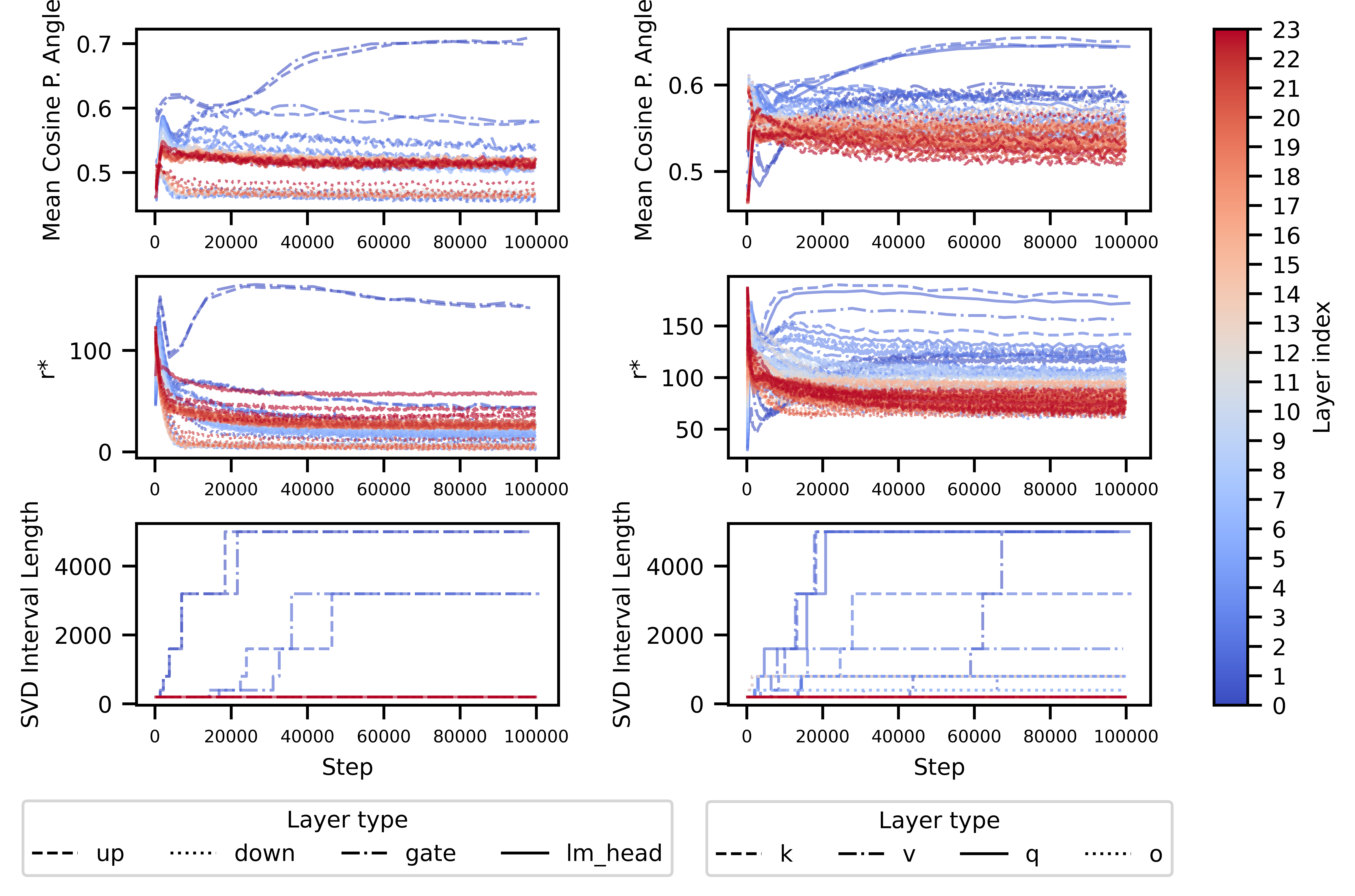}
    \caption{Tracking $r*$ (deterministic rank) and $\rho$ (mean cosine principal angle) during pretraining of Llama 1B on C4 with $\mathrm{PLUMAGE_{S/MP/A}}$}
    \label{fig:alpha_and_r_per_layer}
\end{figure}

\begin{figure}
    \centering
    \includegraphics[width=\linewidth]{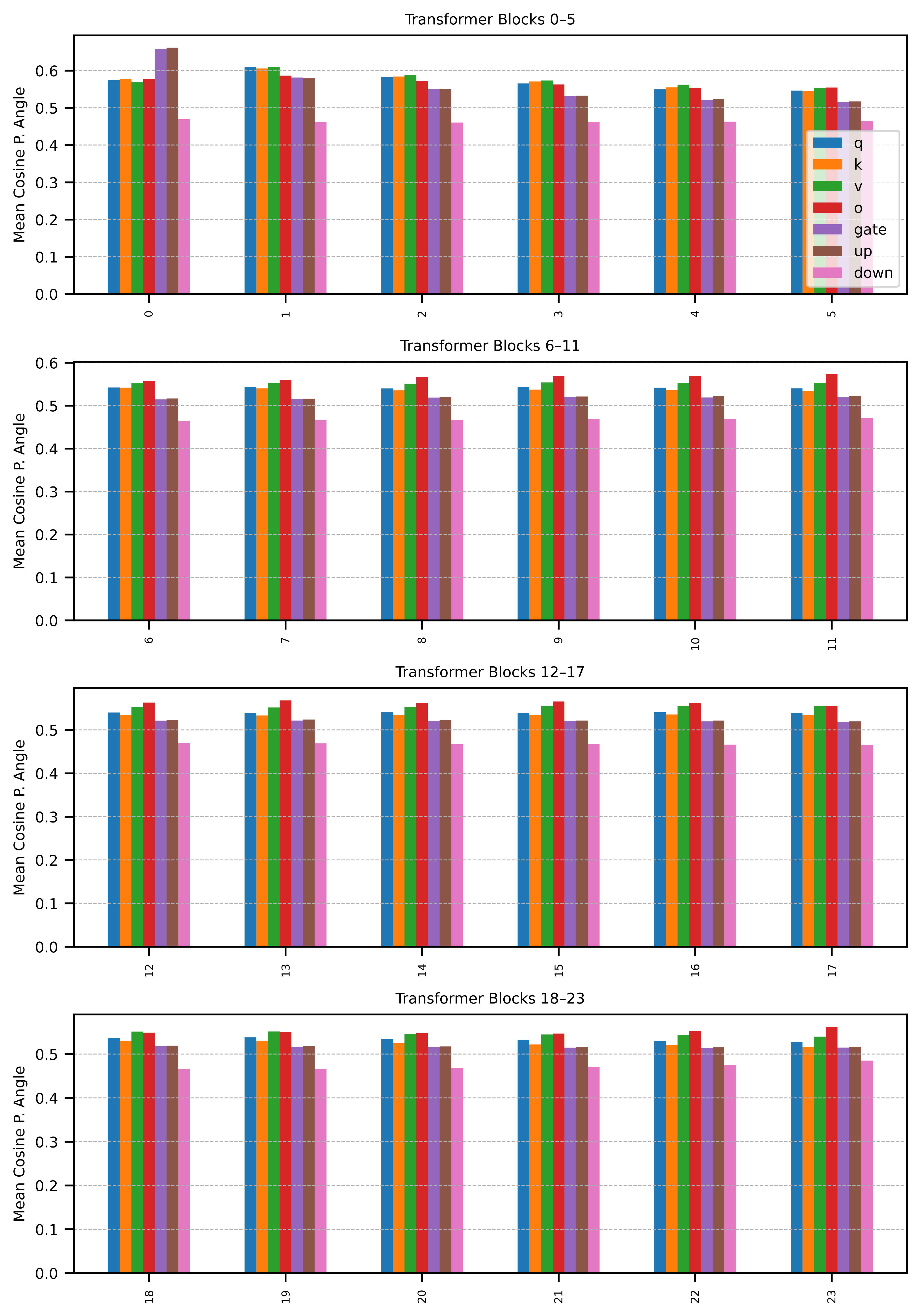}
    \caption{Mean $\rho$ observed during LLama2-1B pretraining on C4 with rank=512}
    \label{fig:mean_cosine_angle_per_layer}
\end{figure}

\begin{figure}
    \centering
    \includegraphics[width=\linewidth]{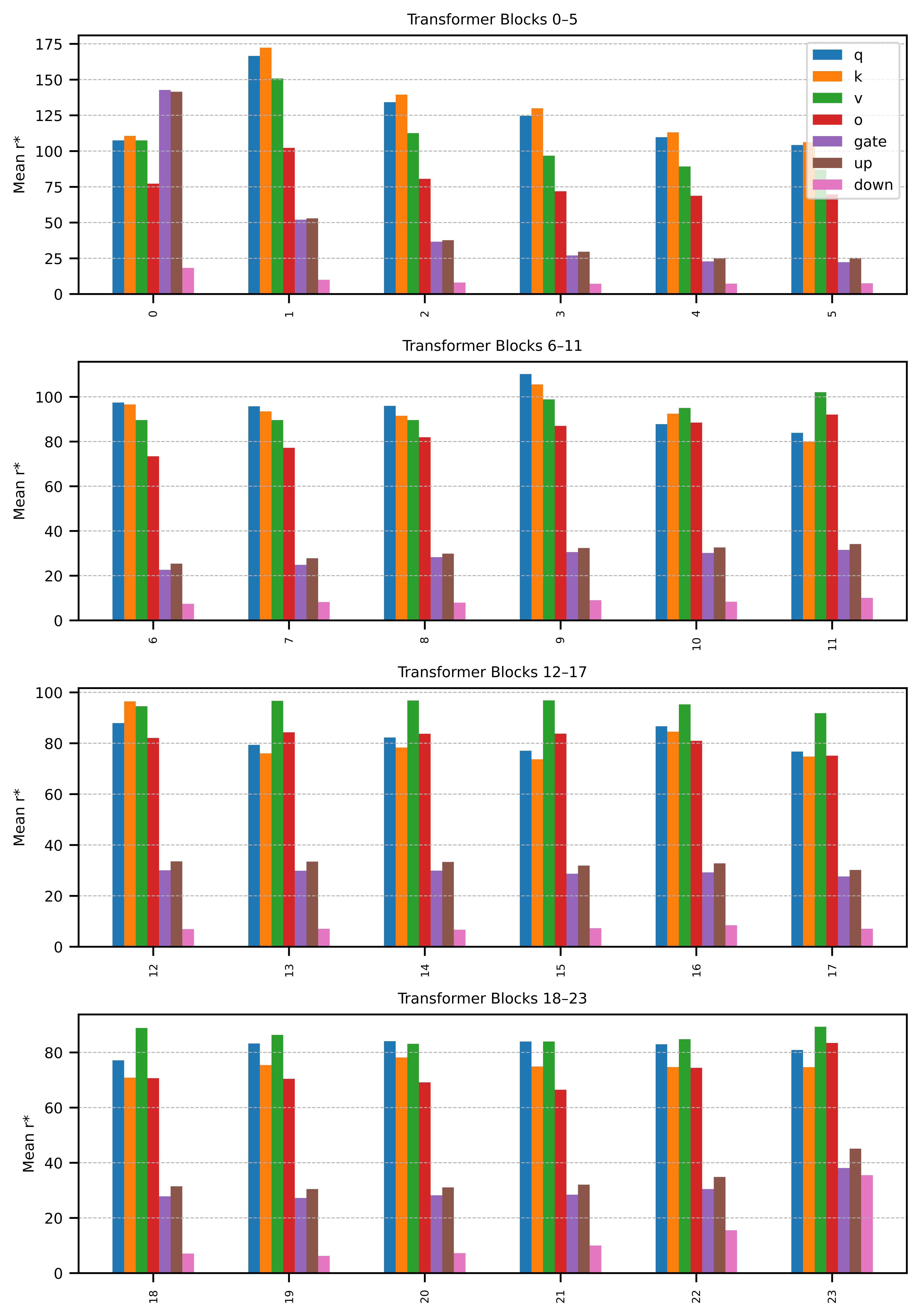}
    \caption{Mean $r^*$ observed during LLama2-1B pretraining on C4 with rank=512}
    \label{fig:mean_r_per_layer}
\end{figure}
\subsection{Low-rank coverage quality}
During training, we monitor the development of $r*$ and $\rho$ for different layers. In \cref{fig:alpha_and_r_per_layer}, we show the progress of the metrics over time, and in (\cref{fig:mean_r_per_layer,fig:mean_cosine_angle_per_layer}), we show the average values per-layer throughout the entire training. By tracking $r^*$ in proportion to the total rank. We find that some layers, such as the attention projection layers, are more suitable for rank reduction. In contrast, the attention MLP down-projection layer parameters are the least amenable. This invites further exploration of rank budget allocation between different layers. For instance, stacking the $QKV$ linear layers into a single matrix allows us to share a single low-rank projection matrix without impacting convergence while reducing \textsc{SVD} computational overhead. Moreover, the memory savings can be translated to increasing the rank of the less amenable layers, such as the MLP down layer, to improve the convergence within the same memory budget. We leave this topic to be explored in future work.

\section{Algorithms}
\subsection{Auxilary algorithms}
In this subsection, we present the algorithms for computing the \textsc{PLUMAGE} probabilities (\cref{alg:getpi}) and sampling from it exactly $k$ indices (\cref{alg:sample_k}) and constructing the projection matrices (\cref{alg:sample_proj}) as discussed in \cref{sec:method,sec:plumage_projections}.
Note that all algorithms can be computed in ($O(\min(m,n))$). This includes the reverse lookup via sorted search in \cref{alg:sample_k} that in case of naive implementation, requires $O(r\mathrm{log}(\min(m,n)))$; however since both arm pointers and target array are sorted, the implementation can continue to iterate over the target array as it locates the arm positions in sub-linear time.
\begin{algorithm}[H]
\caption{\textsc{compute\_sampling\_probabilities}$(\boldsymbol{\sigma},k,\varepsilon)$}
\label{alg:getpi}
\begin{algorithmic}[1]
\STATE {\bfseries Inputs:} $\boldsymbol{\sigma} \in\mathbb{R}^n$ (singular values, descending order), target rank $k$, numerical tolerance $\varepsilon$
\STATE {\bfseries Output:} deterministic rank $r^*$, inclusion probabilities $\boldsymbol{p}\in\mathbb{R}^d$

\STATE $\mathbf{t} \gets \textbf{reverseCumSum}(\boldsymbol{\sigma})$ \RTCOMMENT{$t_i=\sum_{j=i}^d \sigma_j$}
\STATE $\mathbf{t}\gets\max(\mathbf{t},\varepsilon)$ \RTCOMMENT{clip to avoid division by $0$}
\STATE $\mathbf{i}\gets(0,1,\dots,n-1)$ \RTCOMMENT{rank indices}
\STATE $\mathbf{s}\gets k-\mathbf{i}$ \RTCOMMENT{scaling factors $(k-r_t)$}
\STATE $\mathbf{q}\gets (\mathbf{s}\odot \boldsymbol{\sigma})\oslash\mathbf{t}$ \RTCOMMENT{test scores $q_i=\frac{(k-i)\sigma_i}{\sum_{j>i}\sigma_j}$}
\STATE $c\gets\lvert\{i\;|\;q_i<1\}\rvert$ \RTCOMMENT{count how many ranks pass the test}
\STATE $r^*\gets n-c$ \RTCOMMENT{minimal rank that always enters ($r^*=d-c$)}
\STATE $\boldsymbol{p}_{0:r^*-1}\gets 1$ \RTCOMMENT{deterministic inclusion for top $r^*$ modes}
\STATE $\boldsymbol{p}_{r^*:n-1}\gets \dfrac{(k-r^*)\,\boldsymbol{\sigma}_{r^*:n-1}}{t_{r^*}}$ \RTCOMMENT{stochastic inclusion for remaining modes}
\STATE  {\bfseries return:} $r^*,\boldsymbol{p}$
\end{algorithmic}
\end{algorithm}

\begin{algorithm}[H]
\caption{\textsc{sample\_indices}$(\boldsymbol{p},k)$}
\label{alg:sample_k}
\begin{algorithmic}[1]
\STATE {\bfseries Inputs:} inclusion probabilities $\boldsymbol{p}\in\mathbb{R}^n$, target sample size $k$
\STATE {\bfseries Output:} index set $\mathcal{I}\subseteq\{0,\dots , n-1\},\;|\mathcal{I}|=k$

\STATE $\boldsymbol{\tau}\gets\textbf{randPerm}(n)$ \RTCOMMENT{shuffle the indices}
\STATE $\boldsymbol{p}^{\text{sh}}\gets \boldsymbol{p}[\boldsymbol{\tau}]$ \RTCOMMENT{permute the probabilities}
\STATE $\mathbf{c}\gets\textbf{cumSum}(\boldsymbol{p}^{\text{sh}})$ \RTCOMMENT{$c_i=\sum_{j=0}^{i}p^{\text{sh}}_j$}
\STATE $\delta\gets \frac{\sum_{i=1}^{n}p^{\text{sh}}_i}{k}$ \RTCOMMENT{step size = total mass divided by $k$}
\STATE $\beta\gets\textbf{Uniform}(0,\delta)$ \RTCOMMENT{random offset}
\STATE $\mathbf{u}\gets (0,\delta,\dots,(k-1)\delta)+\beta$ \RTCOMMENT{$k$ equally spaced pointers}
\STATE $\mathbf{j}\gets\textbf{searchSorted}(\mathbf{c},\mathbf{u})$ \RTCOMMENT{first index with $c_j\ge \mathbf{u}$}
\STATE $\mathcal{I}\gets \boldsymbol{\tau}[\mathbf{j}]$ \RTCOMMENT{map back to original indices}
\STATE {\bfseries return} $\mathcal{I}$
\end{algorithmic}
\end{algorithm}

\begin{algorithm}[H]
\caption{\textsc{sample\_projections}$(\mathbf{U},\boldsymbol{p},r)$}
\label{alg:sample_proj}
\begin{algorithmic}[1]
\STATE {\bfseries Inputs:} $\mathbf{U} \in\mathbb{R}^{m\times n}$ all singular vectors, column-stacked matrix (assuming $m\leq n$), $\boldsymbol{p} \in\mathbb{R}^n$ sampling probabilities, target rank $r$
\STATE {\bfseries Output:} projections $\mathbf{P} \in\mathbb{R}^{m\times r}$ and $\mathbf{D} \in\mathbb{R}^{r\times r} $
\STATE $\mathcal{I}\leftarrow \text{sample\_indicies}(\boldsymbol{p},r)$
\STATE $\mathbf{P}\leftarrow \mathbf{U}[:,\mathcal{I}]$
\STATE $\mathbf{D}\leftarrow \mathrm{Diag}(\boldsymbol{p}[\mathcal{I}])$
\STATE {\bfseries return} $\mathbf{P},\mathbf{D}$
\end{algorithmic}
\end{algorithm}

\subsection{PLUMAGE full algorithm}
The main plumage algorithm is given in \cref{alg:GaLoRE-v2}. The computational cost is dominated by SVD ($O(d^3)$ assuming gradients are $\mathbb{R}^{d \times d}$ similarly to \textsc{GaLoRE}), which is hard to accelerate in hardware compared to matrix-matrix multiply (GEMM) since it requires sequential matrix-vector products. 
As mentioned in the previous section, the complexity of computing $\boldsymbol{p}$ and sampling from it (\cref{alg:sample_proj},\cref{alg:getpi}) is $O(d)$. We perform both algorithms after each SVD. Thus, the baseline \textsc{PLUMAGE} computational overhead is marginal compared to that of \textsc{GaloRE}. In addition, State update methods ($S/MP$, \cref{eq:MP,v''_approx_A}) are dominated by GEMM operations with complexity $\mathbf{M}^\lfloor: O(d^2r),\mathbf{B}$ and $\mathbf{V}^\lfloor: O(d^2r)$. These operations are easy to accelerate on modern GPUs and are done once per SVD, amortizing their cost. Furthermore, in our experiments using A100/A6000 GPUs, we found that the total training time of our method, using unoptimized implementations, is unaffected by sampling and aligning of moments due to the dominance of SVD overhead. The memory footprint is similar to GaLoRE. It differs by the additional $r$ sampling scale factors per layer. These can be offloaded to the host and prefetched before optimizer weight updates to maintain GaLoRE's memory footprint.
Finally, the adaptive SVD interval method ($S/MP/A$) requires computing the correlation matrix $\mathbf{P}^\top_2\mathbf{P}_1 \in \mathbb{R}^{r^2}: O(r^2d)$ and SVD over the smaller matrix $O(r^3)$. This is done every time the SVD is computed on a new gradient, and the additional SVD step on the small $r\times r$ matrix can be done asynchronously to leverage underutilized compute time since it does not impact the weight update without concurrently storing the old and the fresh projection matrices.

\begin{algorithm}[h]
   \caption{Adam with \textsc{PLUMAGE}}
   \label{alg:GaLoRE-v2}
\begin{algorithmic}
   \STATE {\bfseries inputs:} linear layer weight $\textbf{W} \in \mathbb{R}^{m \times n}$ with $m\leq n$, scalar loss function: $\mathcal{L}:\mathbb{R}^{m \times n}\rightarrow \mathbb{R}^1$, first and second moment decay rates $\beta_1,\beta_2$, learning rate $\eta$, target rank $r$, number of optimization steps $N$, $\tau_0$ number of steps between $\mathrm{SVD}$, $\kappa$ number of steps to resample projection.
   \STATE {\bfseries initialize:} $\textbf{M}_0,\textbf{V}_0 \in \mathbb{R}^{n\times r}$,$\textbf{P}_0 \in \mathbb{R}^{m\times r}\leftarrow I$,$\leftarrow0$, $t\leftarrow0$.
   \REPEAT
   \STATE $\textbf{G}_t\leftarrow \nabla_\textbf{W} \mathcal{L}(\textbf{W}_t)$

   \IF{ $t\mod \tau_t=0$ \OR $t\mod \kappa = 0 $ }
        \IF{ $t\mod \tau_t=0 $ }
            \STATE $\textbf{U}_t,\boldsymbol{\sigma}_t,\textbf{V}_t \leftarrow \mathrm{SVD}(\textbf{G}_t)$
            \STATE $r_t^*,\boldsymbol{p}_t\leftarrow \mathrm{compute\_sampling\_probabilities}(\boldsymbol{\sigma}_t,r,\varepsilon=1e-12)$
        \ELSE
            \STATE $\textbf{U}_t,\boldsymbol{\sigma}_t,\textbf{V}_t \leftarrow \textbf{U}_{t-1},\boldsymbol{\sigma}_{t-1},\textbf{V}_{t-1}$
        \ENDIF
        \STATE $\textbf{P}_t,\textbf{D}_t \leftarrow \mathrm{sample\_projection}(\textbf{U}_t,\boldsymbol{p}_t,r)$ \\\RTCOMMENT{  \cref{alg:sample_proj}}
        \STATE $\textbf{M}_{t},\textbf{V}_{t}\leftarrow \mathrm{update\_state} (\textbf{M}_{t},\textbf{V}_{t},\textbf{P}_t,\textbf{P}_{t-1})$ \\\RTCOMMENT{\cref{eq:MP,v''_approx_A}}
        \STATE  $\tau_{t+1}\leftarrow \mathrm{update\_interval}(\tau_t,\textbf{P}_t,\textbf{P}_{t-1})$ \\\RTCOMMENT{\cref{alg:controller}}
   \ELSE
        \STATE $\mathbf{P}_t,\mathbf{D}_t \leftarrow \mathbf{P}_t,\mathbf{D}_{t-1}$
        
   \ENDIF
   \STATE $\textbf{R}_t\leftarrow \textbf{P}^\top \textbf{G}_t$
   \STATE $\textbf{M}_t\leftarrow  \beta_1 \cdot \textbf{M}_{t-1}+(1-\beta_1) \cdot \textbf{R}_t $
   \STATE $\textbf{V}_t\leftarrow \beta_2 \cdot \textbf{V}_{t-1}+(1-\beta_2) \cdot \textbf{R}_t^{\circ2}$
   \STATE $\textbf{Z}_t\leftarrow \frac{\sqrt{1-\beta_2^t}}{1-\beta_1^{t}}\cdot\frac{\textbf{M}_t}{\sqrt{\textbf{V}_t}+\epsilon}$
   \STATE $\textbf{W}_t\leftarrow \textbf{W}_t - \eta \cdot \textbf{P}_t\mathbf{D}^{-1}\textbf{Z}_t$
   \STATE $t\leftarrow t+1$
   \UNTIL{$t=N$}
\end{algorithmic}
\end{algorithm}
\begin{figure}[h!]
  \centering
  \begin{subfigure}[b]{0.32\linewidth}
    \includegraphics[width=\linewidth]{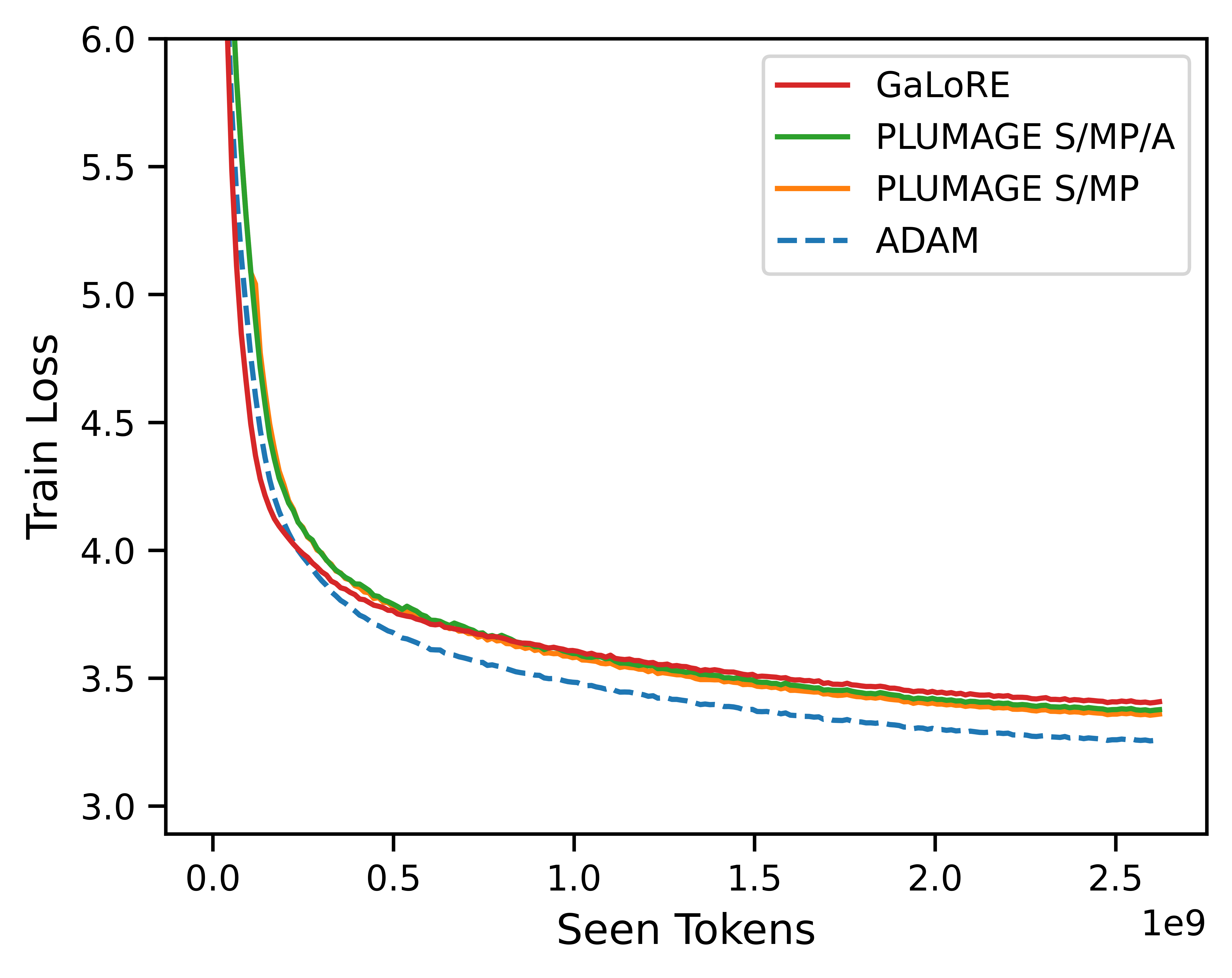}
    \caption{LLama-130M train loss}\label{fig:exp1}
  \end{subfigure}\hfill
  \begin{subfigure}[b]{0.32\linewidth}
    \includegraphics[width=\linewidth]{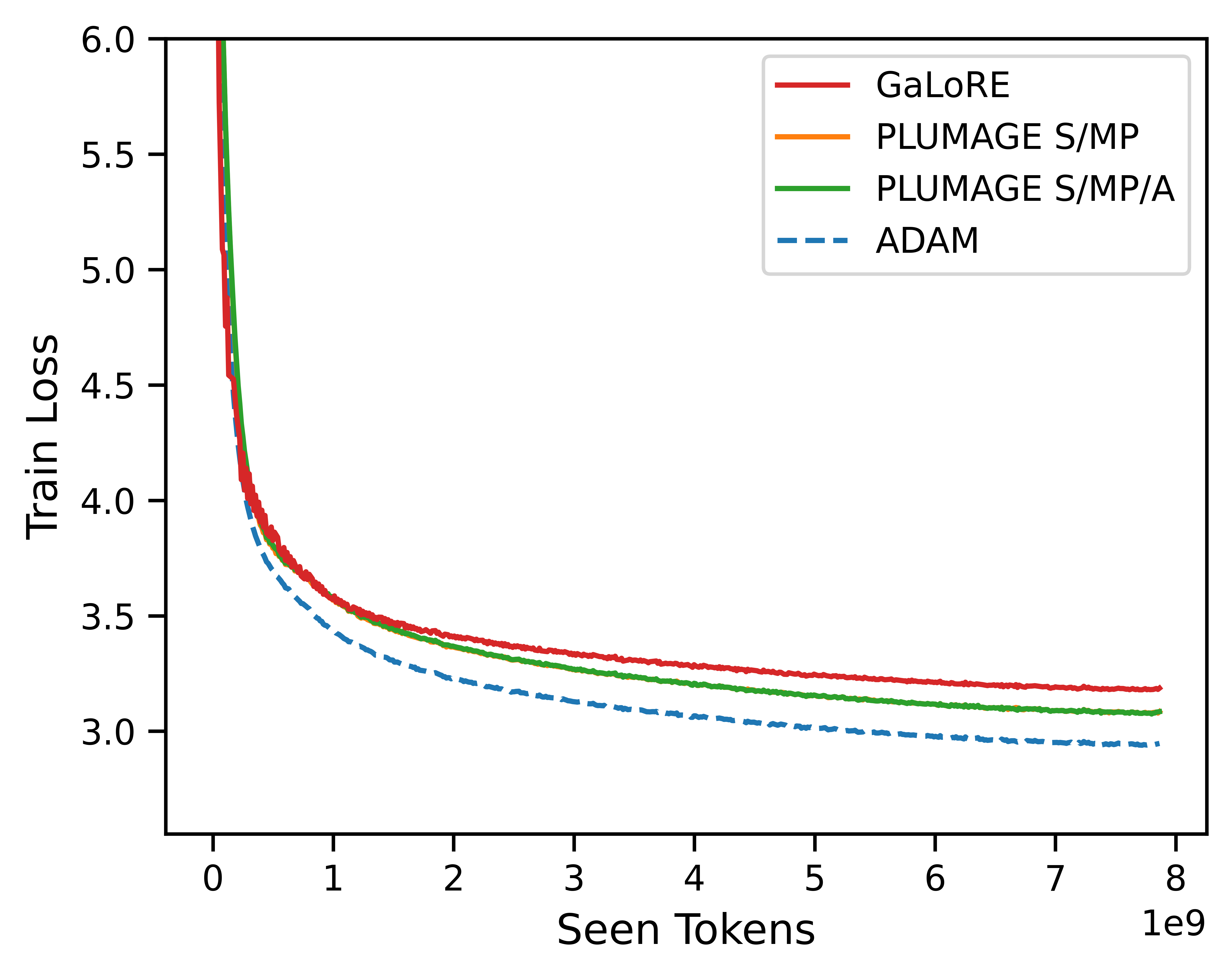}
    \caption{LLama-350M train loss}\label{fig:exp2}
  \end{subfigure}\hfill
  \begin{subfigure}[b]{0.32\linewidth}
    \includegraphics[width=\linewidth]{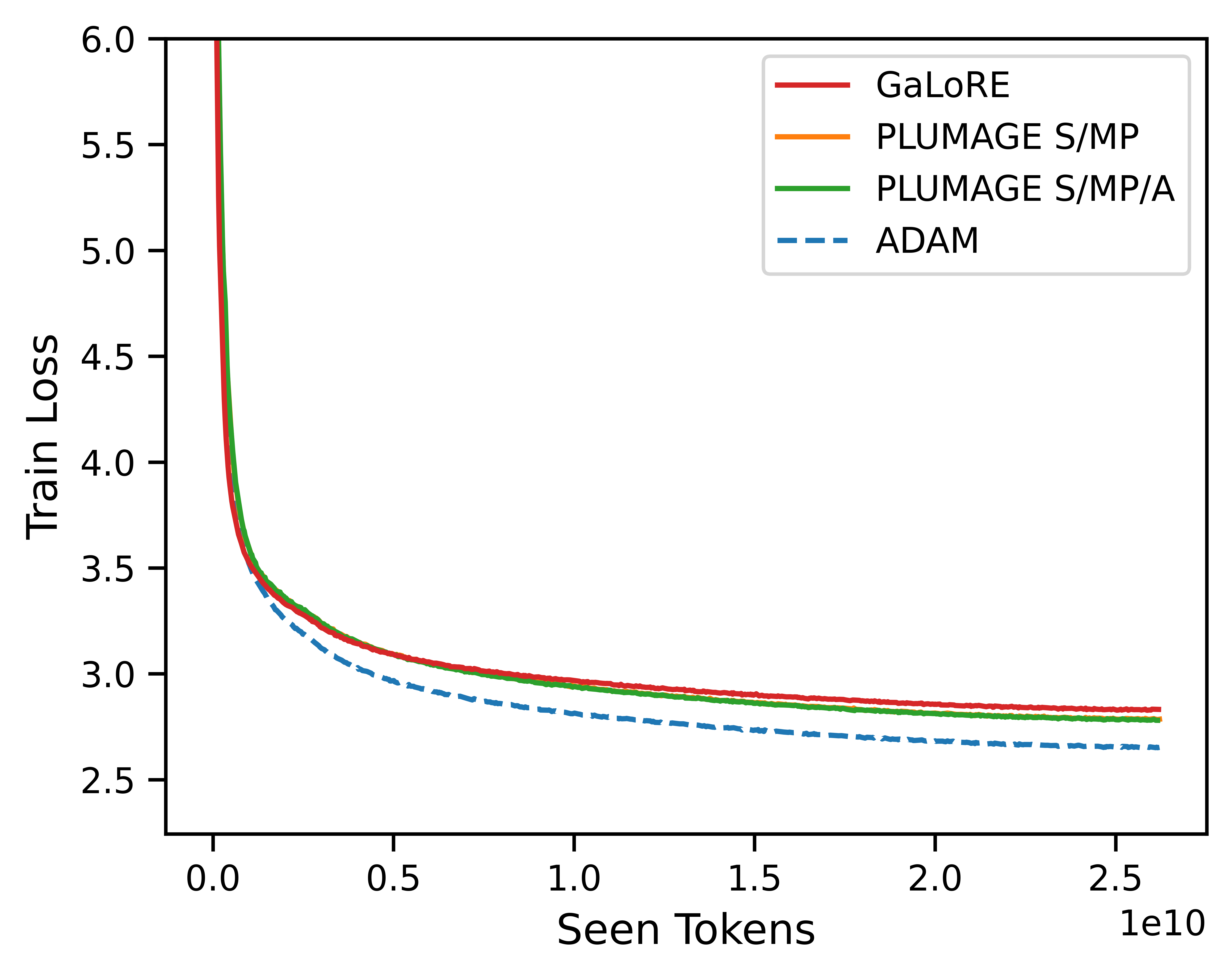}
    \caption{LLama-1B train loss}\label{fig:exp3}
  \end{subfigure}
  \vspace{0.5em} 

  \begin{subfigure}[b]{0.32\linewidth}
    \includegraphics[width=\linewidth]{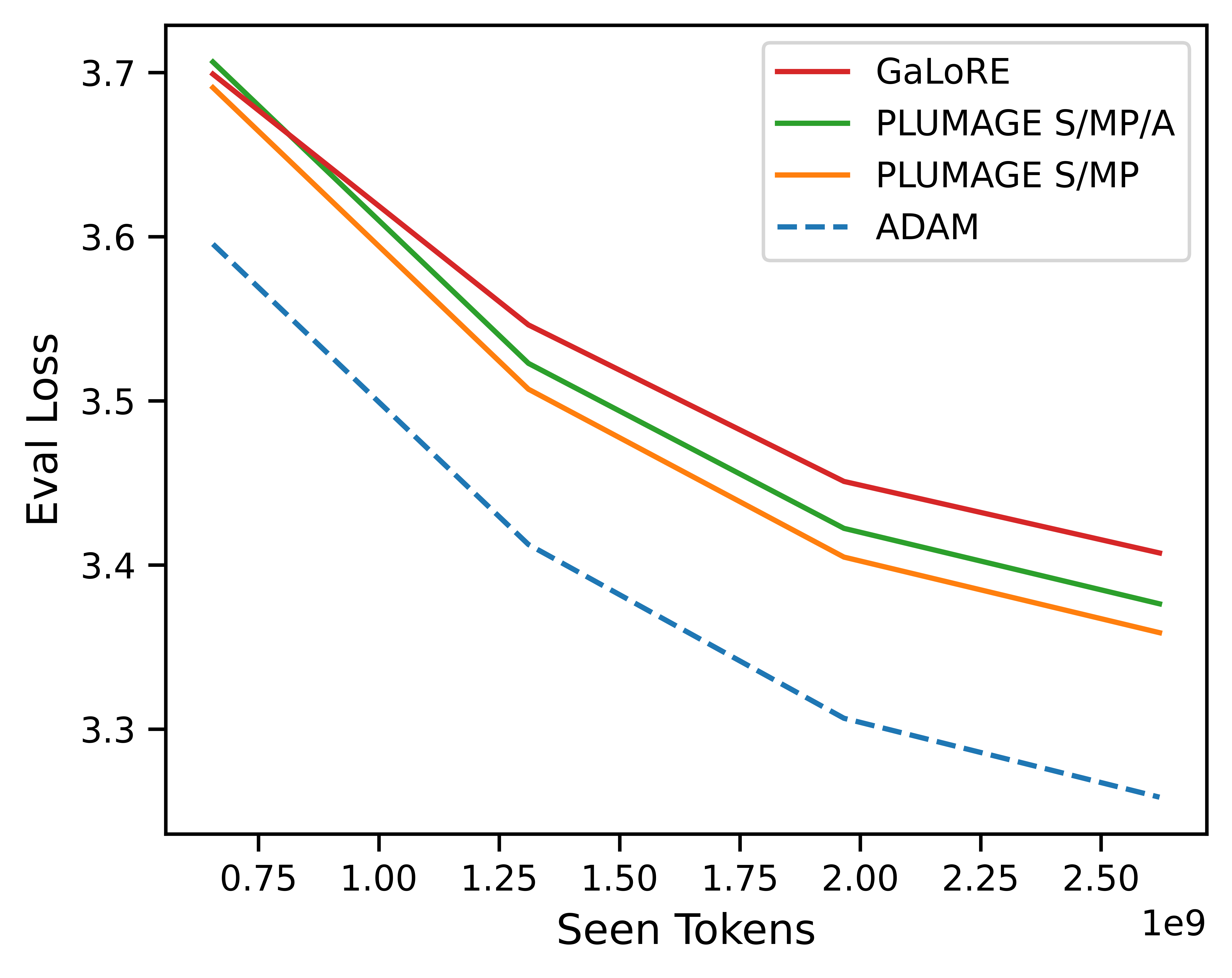}
    \caption{LLama-130M validation loss}\label{fig:exp4}
  \end{subfigure}\hfill
  \begin{subfigure}[b]{0.32\linewidth}
    \includegraphics[width=\linewidth]{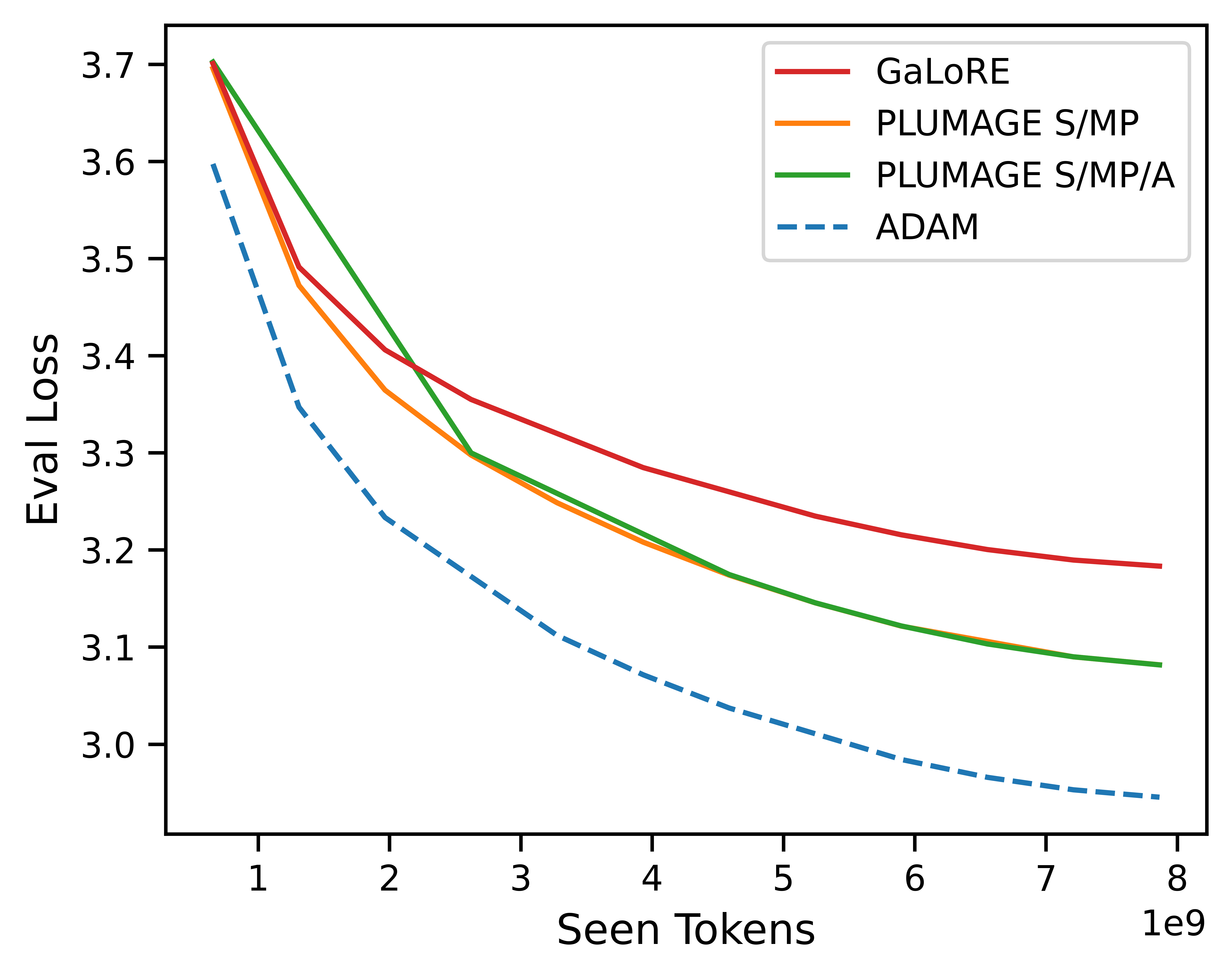}
    \caption{LLama-350M validation loss}\label{fig:exp5}
  \end{subfigure}\hfill
  \begin{subfigure}[b]{0.32\linewidth}
    \includegraphics[width=\linewidth]{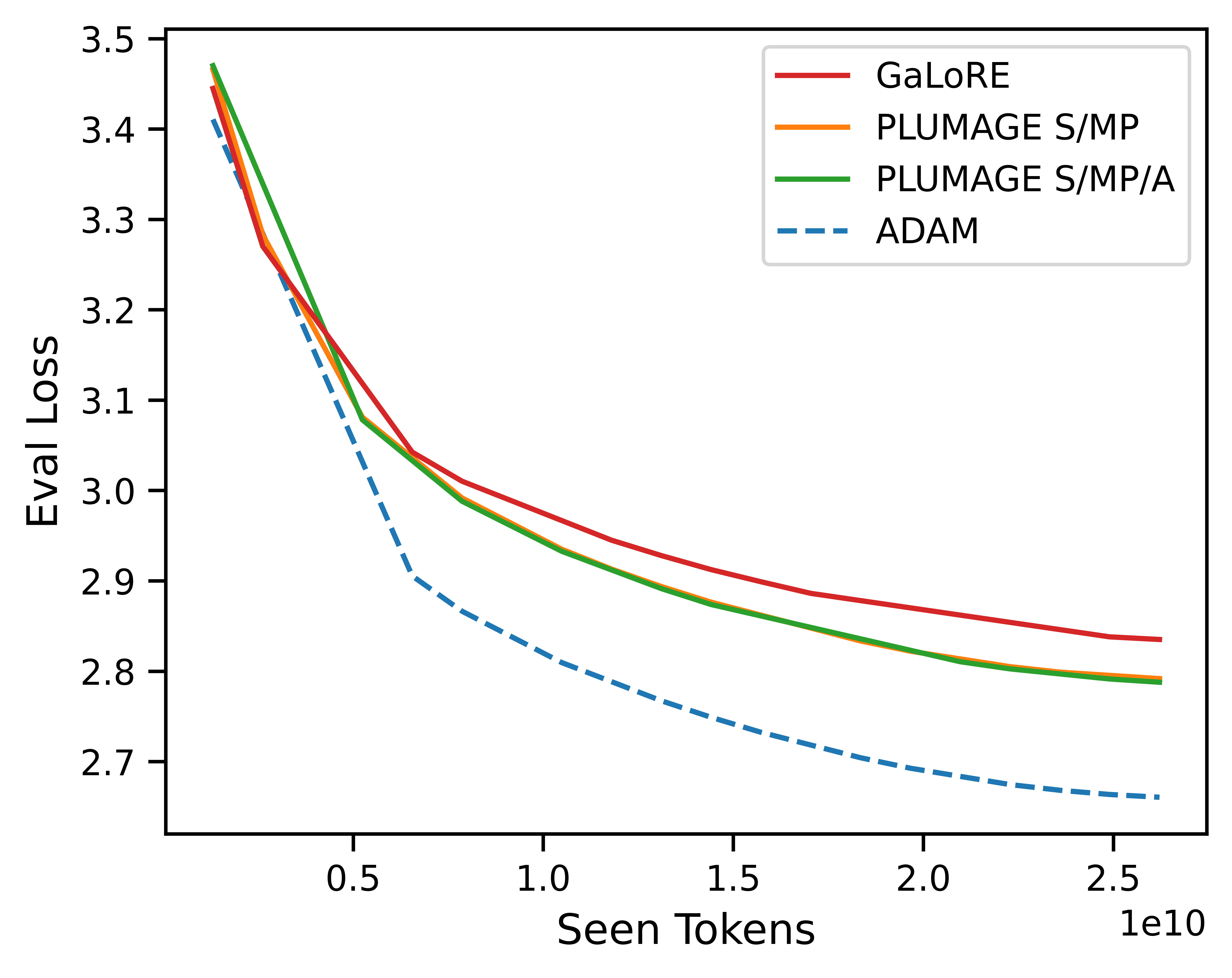}
    \caption{LLama-1B validation loss}\label{fig:exp6}
  \end{subfigure}
  \caption{Pretraining loss plots of Llama on C4 datasts.}
  \label{fig:pretrain_loss}
\end{figure}

\section{Experimental settings}
\label{apdx:exp_settings}
\begin{wraptable}{r}{0.5\textwidth}
    \caption{Model configurations.}
    \label{tab:pretraining_models}
    \resizebox{\linewidth}{!}{%
    \begin{tabular}{c|ccc}
        \toprule
        \textbf{Configuration} & \textbf{130M} & \textbf{350M} & \textbf{1B} \\
        \midrule
        Depth & 12 &24 & 24 \\
        Rank/Hidden &  256/768 & 256/1024 & 512/2048 \\
        Intermidiate & 2048 & 2736 & 5461  \\
        Heads & 12&16& 32 \\
        \bottomrule
    \end{tabular}
    }
    \vspace{-0.5cm}
\end{wraptable}
Our pre-training experiment models are variants of Llama \citep{touvron2023llama2openfoundation}, as suggested by \citet{galore}. These models utilize the same meta-architecture as Llama, except for \textsc{RMS} normalization layers \citep{zhang2019rms} and \textsc{SwiGLU} activation functions \citep{shazeer2020swiglu}. The exact model configuration is given in \cref{tab:pretraining_models}.
In all experiments, we used a single-node GPU server with 8xA100-40GB GPUs or 8xA100-80GB GPUs. The total compute time per experiment varied from $\sim9$ GPU hours for the small 130M parameter model pre-training to $\sim620$ GPU hours for the larger 1B model. We used a slightly modified version of Huggingface transformers \cite{wolf2019huggingfaces} causal model training and GLUE finetuning code examples. We keep the model weights, gradients, and projections in the FP32 data type while using the BF16 mixed-precision support. In \cref{fig:pretrain_loss} we present the full validation and pretraining loss plots for pre-training experiments.

\fi

\end{document}